%% file: main.tex
\documentclass{article}

\usepackage[final]{corl_2026} % Uncomment for the camera-ready ``final'' version.
\usepackage{microtype}
\usepackage{graphicx}
\usepackage{subcaption}
\usepackage{booktabs} % for professional tables
\usepackage{enumitem}
\usepackage{algorithm}

\usepackage{booktabs}
\usepackage{multirow}
\usepackage{amsmath}

\usepackage{amsmath}
\usepackage{amssymb}
\usepackage{mathtools}
\usepackage{amsthm}

\usepackage{algpseudocode}
\usepackage[capitalize,noabbrev]{cleveref}

\theoremstyle{plain}

\theoremstyle{definition}

\theoremstyle{remark}

\usepackage[textsize=tiny]{todonotes}

\title{Learning to Feel the Future: DreamTacVLA for Contact-Rich Manipulation}

\author{
  Guo Ye \textsuperscript{*} \quad
  Zexi Zhang \textsuperscript{*} \quad
  Xu Zhao \quad
  Shang Wu\quad
  Haoran Lu\quad
  Shihan Lu \textsuperscript{\ensuremath{\dagger}} \quad
  Han Liu \textsuperscript{\ensuremath{\dagger}}\\
  {\normalfont \textsuperscript{*}Equal Contribution} \quad   \textsuperscript{\ensuremath{\dagger}}Equal Advising\\
  {\normalfont Northwestern University}\\
  \texttt{guoye2018@u.northwestern.edu},  
  \texttt{\{shihanlu,hanliu\}@northwestern.edu}\\
}

\begin{document}
\maketitle

%===============================================================================

\begin{abstract}
Vision-Language-Action (VLA) models have shown remarkable capability %generalization
in mapping web-scale knowledge to robotic control, yet they remain largely blind to physical contact. This limits their performance on contact-rich manipulation tasks, where successful execution requires not only perceiving contact but also anticipating how contact evolves during interaction. 
% As a result, they struggle with contact-rich manipulation tasks that require reasoning about interactions. %force, texture, and slip. 
% A key limitation is their inability to anticipate how tactile signals will evolve during interaction, which is essential for %reasoning about contact dynamics and 
% guiding precise manipulation. 
%Existing approaches that incorporate low-dimensional tactile signals still fail to capture the high-resolution contact dynamics essential for such interactions. 
% While some approaches incorporate low-dimensional tactile signals, they fail to capture the high-resolution dynamics essential for such interactions. 
To address this limitation, we introduce \textbf{DreamTacVLA}, a framework that grounds VLA models in contact interactions by learning to ``feel the future.'' 
DreamTacVLA adopts a hierarchical perception scheme that integrates tactile signals with wrist-camera and third-person views. We first train a unified policy with a Hierarchical Spatial Alignment (HSA) loss to align these multi-scale modalities and generate a draft action. 
% DreamTacVLA first uses a hierarchical perception scheme combining tactile signals with wrist and third-person views, and trains a unified policy with a Hierarchical Spatial Alignment (HSA) loss to align these modalities and generates a draft action. 
% DreamTacVLA adopts a hierarchical perception scheme that treats high-resolution tactile images as micro-vision inputs, complemented by wrist-camera local vision and third-person macro vision. To integrate these multi-scale sensory streams, we first train a unified policy with a Hierarchical Spatial Alignment (HSA) loss, which aligns tactile tokens with their corresponding spatial features in the wrist and third-person views, and generate a draft action. %To further deepen the model’s understanding of fine-grained contact dynamics, 
This draft action is then combined with future tactile embeddings predicted by a pretrained tactile world model to produce a contact-prediction latent, which anticipates the contact consequences of its own action.
% The draft action, with predicted tactile observations from a pretrained tactile world model, then produces a latent that anticipates the contact consequences of its own action. %During fine-tuning, 
% Together with the current observation, this contact-prediction latent is used to generate the final action for execution. 
Conditioned on both the current observation and this contact-prediction latent, DreamTacVLA generates the final action for execution. 
% This latent is used to generate the final action for execution, conditioning the policy on both current observations and the anticipated contact consequences.  
% these predicted tactile states are used to refine the policy, enabling DreamTacVLA to produce a final action for execution that is conditioned on both current observations and anticipated contact states.  
% We then finetune the system with a tactile world model that predicts future tactile observations, enabling the policy to reason about fine-grained contact dynamics before taking action. 
Additionally, to mitigate tactile data scarcity and the wear-prone nature of tactile sensors, we construct a hybrid large-scale dataset from both high-fidelity digital twin and real-world experiments. 
% By anticipating future tactile states, DreamTacVLA learns a rich representation of contact interactions and generates actions based on both current observations and imagined consequences. 
Across four contact-rich manipulation tasks, DreamTacVLA outperforms state-of-the-art VLA and imitation-learning baselines with and without tactile sensing, achieving up to 95\% success. These results highlight the importance of action-conditioned tactile prediction for contact-rich robotic manipulation. 
%understanding physical contact for robust, touch-aware robots. %robotic agents. 
Code, data, and videos are available at~\url{https://michaelyeah7.github.io/learning-to-feel-the-future/}.

\end{abstract}

\keywords{Tactile Sensing, World Models, Contact-Rich Manipulation} 

%===============================================================================

\input{sec/1_intro}
\input{sec/2_relatedwork}

\input{sec/3_method}

\input{sec/4_experiments}

\input{sec/5_conclusion}

%===============================================================================

\clearpage
% The acknowledgments are automatically included only in the final and preprint versions of the paper.
% \acknowledgments{If a paper is accepted, the final camera-ready version will (and probably should) include acknowledgments. All acknowledgments go at the end of the paper, including thanks to reviewers who gave useful comments, to colleagues who contributed to the ideas, and to funding agencies and corporate sponsors that provided financial support.}

%===============================================================================

% \bibliography{main}  % .bib
\bibliography{main2}  % .bib

\newpage
\appendix
\onecolumn

\input{sec/6_supplementary}

\end{document}

%% file: sec/1_intro.tex
\section{Introduction}
\label{sec:intro}

\begin{figure*}
    \centering
    \includegraphics[width=0.92\linewidth]{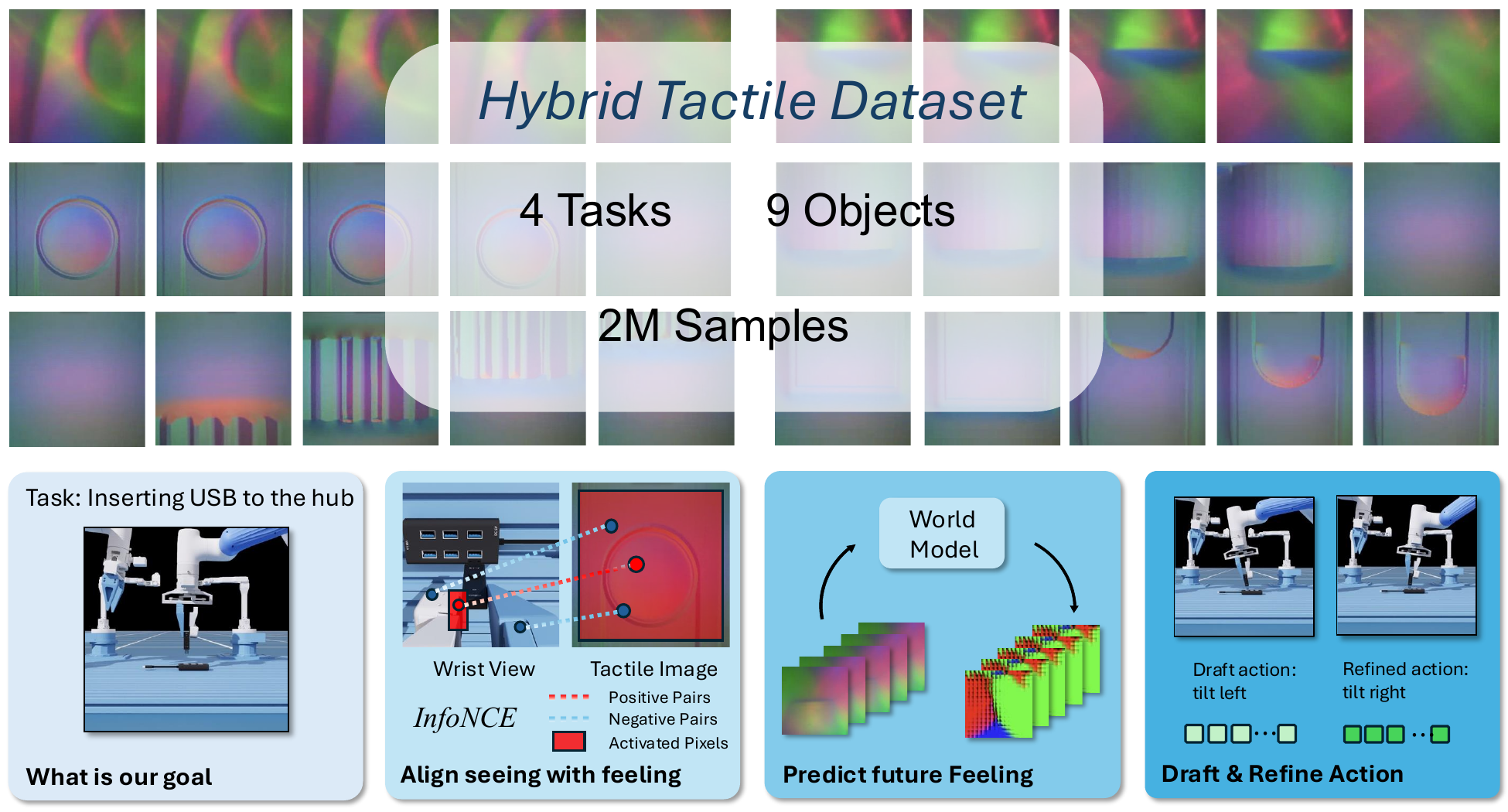}
    % \hspace{-1.0cm}
    \caption{Hybrid tactile dataset and DreamTacVLA inference pipeline. (Top) We collect a large-scale tactile dataset spanning four manipulation tasks and nine objects, comprising 2 million tactile frames. (Bottom) Our Think–Dream–Act employs a two-pass inference process. In the \textbf{Think} stage, the policy proposes a draft action based on the current observation and a null tactile prediction. In the \textbf{Dream} stage, a frozen V-JEPA 2 world model forecasts the tactile outcome of the draft action. In the \textbf{Act} stage, the policy integrates both the current observation and predicted tactile outcome to refine the action, enabling fine-grained corrections during contact-rich manipulation.}
    \label{fig:think-dream-act}
    % \vspace{-1em}
\end{figure*}

Vision-Language-Action (VLA) models leverage web-scale knowledge for general-purpose robot manipulation \cite{Ghosh-RSS-24, pmlr-v270-kim25c, pmlr-v229-zitkovich23a}, but remain largely limited to visually guided tasks. In contact-rich scenarios such as insertion, % or deformable object manipulation, 
their lack of tactile awareness often leads to failure. Although recent work has introduced tactile inputs into VLAs~\cite{huang2025tactilevla, cheng2025omnivtla}, these approaches rely on sparse, ambiguous force or torque signals, providing limited information about contact location and dynamics.

% Vision-Language-Action (VLA) models enable robots to leverage web-scale knowledge for general-purpose manipulation \cite{Ghosh-RSS-24, pmlr-v270-kim25c, pmlr-v229-zitkovich23a}, but their success is largely limited to visually guided tasks. In contact-rich scenarios such as insertion or deformable object manipulation, VLA agents often fail due to the lack of tactile awareness. Although recent work has introduced tactile inputs into VLA pipelines \cite{huang2025tactilevla, cheng2025omnivtla}, these approaches rely on low-dimensional force or torque signals that are sparse and ambiguous, providing little information about how or where contact occurs.

High-resolution tactile sensing provides rich contact information, but sensor cost and fragility limit data scalability. To address this bottleneck, we develop a large-scale simulation pipeline with a high-fidelity digital twin of the tactile sensor and manipulation environment, enabling scalable data generation and improved sim-to-real transfer. 
We further structure perception hierarchically across three spatial scales to reason about contact interactions: \textbf{macro} third-person task context, \textbf{local} end-effector visual guidance, and \textbf{micro} fingertip tactile feedback. 
%To reason over contact-rich interactions, we further organize perception hierarchically across three spatial scales: \textbf{macro} third-person-view task context, \textbf{local} end-effector visual guidance, and \textbf{micro} fingertip tactile feedback.

% To create robots capable of human-level dexterity, tactile sensing must be high-resolution and integrated across multiple spatial scales. However, scaling such tactile-aware models poses a fundamental data challenge: visual tactile sensors are expensive and fragile, making large-scale real-world data collection prohibitively costly. We therefore construct a large-scale tactile data generation pipeline in simulation, complemented by a high-fidelity digital twin of the tactile sensor and manipulation environment to improve sim-to-real transfer. This hybrid strategy enables scalable tactile learning while maintaining physical realism.

Because vision and touch differ substantially in representation and semantics, directly combining them may not enable effective multimodal reasoning. We use robot kinematics and camera calibration to establish spatial correspondence between tactile inputs and wrist and third-person views, learning a unified latent representation across modalities. %We establish spatial correspondence between tactile inputs and their locations in wrist and third-person views using robot kinematics and camera calibration, and learn a unified latent representation across the aligned modalities. 
This alignment is particularly important for vision-language backbones pretrained without touch, which may otherwise ignore tactile inputs despite their unique ability to capture fine-grained contact phenomena, such as slip and local deformation.

% However, data alone are not sufficient. Contact-rich manipulation inherently requires reasoning across multiple spatial scales, from global task context to fine-grained contact events. This motivates our hierarchical perception framework, which organizes sensory inputs into three levels: \textbf{macro} for arm-level task context, \textbf{local} for end-effector visual guidance, and \textbf{micro} for fingertip tactile cues such as slip and insertion forces (Figure~\ref{fig:three-scale}).

To further encourage tactile reasoning, we introduce a tactile world model that predicts future tactile states in latent space. Unlike conventional world models that predict high-dimensional visual observations, tactile signals exhibit more constrained dynamics while directly encoding contact interactions, making them well suited for predictive modeling. 
% Language provides lightweight task-level conditioning, while our focus remains on tactile-visual reasoning and predictive modeling of contact interactions. 
Rather than relying on %this model with 
reward-based planning or MPC, we propose DreamTacVLA, using a two-stage framework: the policy first generates a draft action from aligned multi-scale observations, with language providing lightweight task-level conditioning, and then refines the action using predicted future tactile latent states. This design emphasizes tactile-visual reasoning, enabling the policy to jointly interpret current sensory observations and anticipate how contact interactions will evolve. %In this way, DreamTacVLA jointly reasons over what the robot currently sees and feels and how contact will evolve.

In summary, we make the following contributions:
\begin{itemize} [leftmargin=2em, itemsep=0.08em, topsep=0.15em]
     \item A multi-scale contrastive loss that spatially aligns heterogeneous sensor modalities in a unified latent space. %We introduce a novel contrastive loss for spatial alignment on multi-scale sensor data. This method aligns diverse perceptions into a single unified latent space.
    \item A self-supervised tactile world model that predicts future tactile latent, implicitly capturing contact transitions and interactions. %We introduce a tactile world model trained as a self-supervised objective to “dream" the future. By predicting high-resolution tactile signals, this model learns an implicit understanding of contact physics and object interactions.
    \item A two-stage ``Think-Dream-Act'' framework that first predicts a draft action, then forecasts its tactile consequences, and finally refines the action accordingly. %We propose a two-stage “Think-Dream-Act" policy that uses this “dreaming" capability for action refinement. The policy first thinks of a draft action, then dreams its tactile consequences using the world model, and finally acts by outputting a refined, more precise command.
    % \item We introduce a large-scale simulated tactile dataset, paired with a high-fidelity digital twin, which enables dense and diverse tactile supervision that would be prohibitively expensive to obtain in the real world.

    % \item We introduce a novel contrastive loss for spatial alignment on multi-scale sensor data. This method aligns diverse perceptions into a single unified latent space.
    % \item We introduce a tactile world model trained as a self-supervised objective to “dream" the future. By predicting high-resolution tactile signals, this model learns an implicit understanding of contact physics and material interactions.
    % \item We propose a two-stage “Think-Dream-Act" policy that uses this “dreaming" capability for refinement. The policy first thinks of a draft action, then dreams its tactile consequences using the world model, and finally acts by outputting a refined, more precise command.
    % \item We introduce a large-scale simulated tactile dataset, paired with a high-fidelity digital twin, which enables dense and diverse tactile supervision that would be prohibitively expensive to obtain in the real world.
\end{itemize}

%% file: sec/2_relatedwork.tex
\section{Related Work}
\label{sec:related_work}
\subsection{Vision-Language-Action (VLA) Models}
% Vision-Language-Action (VLA) models have become a standard paradigm for general-purpose robot control, with large-scale systems that demonstrate strong generalization of the tasks and the embodiments \cite{brohan2022rt, brohan2024rt, chi2025diffusion, zhao2023learning, o2024open, team2024octo, kim2024openvla, liu2024rdt, black2024pi_0, bjorck2025gr00t, intelligence2025pi_}. Although these works scale model capacity, action representations, and dataset diversity, they remain predominantly vision-centric and continue to struggle in contact-rich manipulation where tactile reasoning is essential, motivating recent attempts such as FuSe \cite{jones2025beyond} to finetune generalist policies with touch (and audio).

% Beyond generalist control, CogACT \cite{li2024cogact} introduce a cognitively structured, componentized VLA architecture that decouples high-level reasoning from fine-grained action synthesis. Most recently, optimized fine-tuning strategies, such as OpenVLA-OFT \cite{kim2025fine} show that carefully designed adaptation recipes can substantially boost real-robot success rates and inference efficiency. 
Vision-Language-Action (VLA) models have become a dominant paradigm for general-purpose robot control, demonstrating strong cross-task and cross-embodiment generalization by scaling data, model capacity, and action representations \cite{Brohan-RSS-23, pmlr-v229-zitkovich23a, Ghosh-RSS-24, pmlr-v270-kim25c, black2024pi_0}. Recent advances also improve VLA effectiveness via modular architectures and stronger adaptation recipes, boosting real-robot performance and efficiency \cite{li2024cogact, kim2025fine}. Despite these advances, most VLAs remain predominantly vision-centric and continue to struggle in contact-rich manipulation where tactile reasoning is critical, motivating efforts to incorporate touch into generalist policies via post-hoc adaptation~\cite{11127987}. %Detailed discussion and additional references please refer to Appendix~\ref{appRel1}.

\subsection{Multimodal Grounding for Robotics}
Multimodal grounding aligns perception with physical interaction by incorporating spatial and tactile cues beyond RGB. Spatially enhanced VLAs introduce explicit or implicit geometric priors (e.g., egocentric 3D encodings, point clouds, or structured traces) to improve geometric reasoning and manipulation robustness \cite{QuD-RSS-25, 11346992, lin2025evo}. Tactile grounding further augments contact awareness, yet many policies still rely on low-dimensional force or compressed touch signals that miss fine-grained contact geometry; recent work instead fuses high-resolution tactile sensing for multimodal reasoning and reactive correction in contact-rich tasks \cite{liu2025mla, huang2025tactilevla, cheng2025omnivtla, XueH-RSS-25}. In parallel, tactile representation learning targets transferable visuotactile embeddings via self-supervision or cross-modal alignment \cite{yang2024binding, pmlr-v235-fu24b, pmlr-v270-higuera25a}. Our work leverages vision-based tactile micro-vision to capture texture, geometry, and shear-induced slip for fine-grained contact modeling. %Detailed discussion and additional references please refer to Appendix~\ref{appRel2} and \ref{appRel3}.

\subsection{Predictive World Models in Robotics}
% Classical latent-dynamics frameworks such as World Models \cite{ha2018world} showed that generative models can learn predictive environment dynamics. Dreamer \cite{hafner2023mastering} improved temporal coherence and scaled such models to long-horizon tasks through richer latent dynamics. More recent approaches \cite{assran2025v, nair2022r3m} leverage large-scale pretraining to produce visual encoders whose structured latent spaces capture temporal, physical, and semantic regularities beneficial for downstream control. 

% In the tactile domain, ViTacFormer \cite{heng2025vitacformer} learns visuo-tactile representations with an autoregressive tactile prediction objective. Recent VLA architectures \cite{zhang2025dreamvla, cen2025worldvla} integrate predictive world models directly into the policy, enabling action generation to be conditioned on latent rollouts that capture future scene evolution. 

% Our work generalizes this paradigm to the tactile domain, predicting high-resolution tactile futures and aligning them with visual observations through a hierarchical spatial mechanism.
Predictive world models learn latent dynamics that capture temporal and physical regularities, enabling agents to anticipate future evolution for decision making \cite{ha2018world, hafner2023mastering}. Large-scale pretraining further yields structured representations whose latent spaces encode semantics and dynamics beneficial for downstream control~\cite{pmlr-v205-nair23a, assran2025v}. Recent VLA architectures integrate such predictive modeling into policy learning by conditioning actions on latent rollouts \cite{NEURIPS2025_22d4f952, cen2025worldvla}. In the tactile domain, prior work mostly studies representation learning or autoregressive tactile prediction \cite{heng2025vitacformer}, whereas our work predicts high-resolution tactile futures and aligns them with visual observations for fine-grained contact reasoning. More discussion and references on related work can be found in Appendix~\ref{appendix:extended_related_work}.

%% file: sec/3_method.tex
\section{Methodology}
\begin{figure*}[t]
    \centering
    \includegraphics[width=1.0\linewidth]{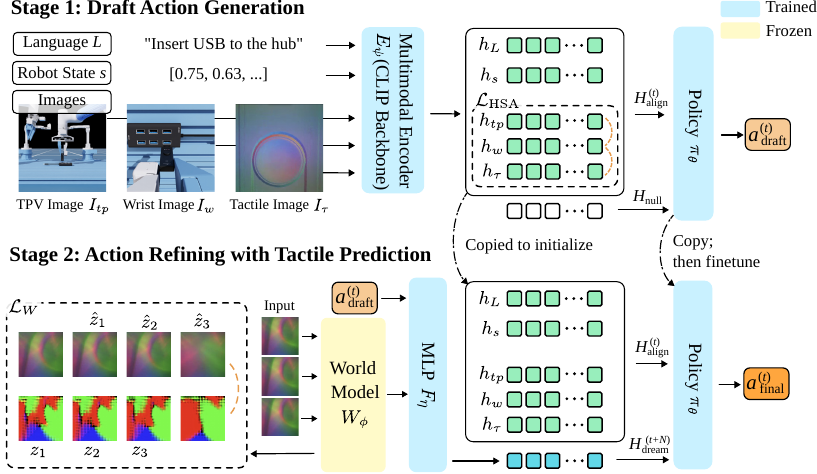}
    % \caption{The DreamTacVLA architecture. A unified policy $\pi_{\theta}$ and a multimodal world model $W_{\phi}$ are trained end-to-end. The policy first generates a draft action, which the world model uses to dream the future tactile and visual state. The policy then uses this dream to refine its action, grounding its decision in predicted contact physics.}
    \caption{The proposed framework operates in two stages. \textbf{Stage 1:} A multimodal encoder $E_{\psi}$ processes diverse inputs. This stage employs Hierarchical Spatial Alignment (HSA) to effectively fuse the features from different modalities, guided by the $\mathcal{L}_\mathrm{HSA}$ and $\mathcal{L}_W$ losses. A policy $\pi_{\theta}$ is trained to output an initial draft action $a^{(t)}_{\mathrm{draft}}$. \textbf{Stage 2:} A world model $W_{\phi}$ is trained to predict future tactile image sequences. The policy ``dreams'' the future tactile feeling (e.g., $H^{(t+N)}_{\mathrm{dream}}$) that would result from its draft action. This predicted future is fed into an MLP, allowing the policy to refine its plan and output a more refined final action $a^{(t)}_{\mathrm{final}}$.}
    \label{fig:architecture}
% \vspace{-1.5em}
\end{figure*}

Our model, DreamTacVLA, learns robust contact-rich manipulation by integrating high-resolution vision-based tactile images with standard visual (third-person and wrist camera) and language inputs. As shown in Figure~\ref{fig:architecture}, it is a unified, end-to-end framework built on a shared LLM backbone with three key components. \textbf{Multimodal Encoders ($E_{\psi}$)}: modality-specific encoders (a CLIP ViT for third-person/wrist images and language, an MLP for robot state) produce feature tokens concatenated into a unified sequence; in Stage 1 the vision and tactile encoders are trained with our Hierarchical Spatial Alignment (HSA) loss, yielding a spatially-aligned representation $H_\mathrm{align}^{(t)}$. \textbf{Tactile World Model ($W_{\phi}$)}: an implicit model that takes the current tactile images and a draft action $a_\mathrm{draft}^{(t)}$ to predict the future tactile latent $H_\mathrm{dream}^{(t+N)}$. \textbf{Unified Policy ($\pi_{\theta}$)}: a CLIP-based~\cite{pmlr-v139-radford21a} encoder paired with a transformer-based action expert that operates in two passes—first drafting $a_\mathrm{draft}^{(t)}$ from the current observations, then producing a refined action $a_\mathrm{final}^{(t)}$ from both $H_\mathrm{align}^{(t)}$ and the dreamed future tactile latent $H_\mathrm{dream}^{(t+N)}$. This Think–Dream–Act loop %, implemented through two-stage training, lets
allows the policy internally verify decisions by forecasting the contact consequences of the draft action before execution.

\subsection{Stage 1: Draft Action Generation} %Pre-training Spatial Alignment \& World Model}
In this stage, we train the multimodal encoders and an initial action policy to learn spatially grounded representations of visual and tactile observations.
% we train the observation encoders to capture the spatial relationship between tactile sensing and the visual environment while learning an initial action policy. 
This is achieved by jointly optimizing the action loss ($\mathcal{L}_{\mathrm{action}}$) and our proposed Hierarchical Spatial Alignment (HSA) loss ($\mathcal{L}_{\mathrm{HSA}}$). Separately, we pretrain a tactile world model ($W_{\phi}$) on unlabeled tactile sequences and freeze it for %subsequently 
use in the next stage.

\subsubsection{Hierarchical Spatial Alignment (HSA)}

To establish spatial correspondence across the three sensing scales, we introduce a Hierarchical Spatial Alignment (HSA) loss that aligns tactile representations with their corresponding regions in wrist and third-person camera views. 

% To enable the model to fuse information across the three visual scales (third-person view, wrist, and tactile image), it must understand where the tactile sensor is located within the other camera views. We promote this understanding through a Hierarchical Spatial Alignment (HSA) loss.

% The three-scale visual hierarchy of our model. Our framework fuses information from three distinct visual modalities: (a) The 'Macro Vision' (Third-Person) camera, which captures the entire workspace. (b) The 'Local Vision' (Wrist) camera, which provides a close-up, egocentric view of the end-effector and the target object. (c) The 'Micro Vision' (Tactile) sensor, which outputs a high-resolution image of the physical contact surface itself. Our Hierarchical Spatial Alignment (HSA) loss is designed to explicitly ground the 'micro-vision' (what the robot feels) within the 'local' and 'macro' visual contexts (what the robot sees).

First, using the robot's forward kinematics and calibrated camera parameters (extrinsics $E_{tp}, E_w$ and intrinsics $K_{tp}, K_w$), we find the 3D pose of the tactile sensor $P_\mathrm{sensor}^{(t)} \in SE(3)$. We then project this pose to find its corresponding 2D bounding box in both camera views: $\mathcal{B}_{w}^{(t)}$ in the wrist view and $\mathcal{B}_{tp}^{(t)}$ in the third-person view.

% From an intermediate layer of the LLM, we extract the feature tokens $H_{mid}^{(t)}$. 
We compute three mean-pooled feature vectors:
(1) $h_\tau$: The mean-pooled embedding of all tactile tokens $Z_\tau^{(t)}$.
(2) $h_w$: The mean-pooled embedding of all wrist-view tokens whose spatial positions fall within the projected bounding box $\mathcal{B}_w^{(t)}$.
(3) $h_{tp}$: The mean-pooled embedding of all third-person view tokens within $\mathcal{B}_{tp}^{(t)}$.

\begin{figure}[tb]
    \centering    \includegraphics[width=0.8\linewidth]{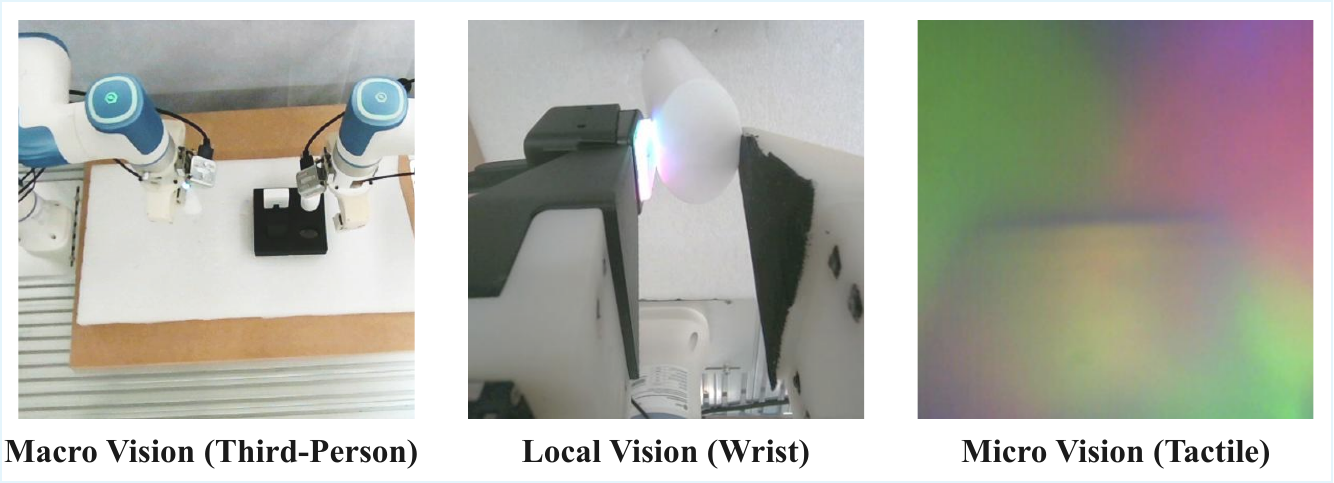}
    \caption{The three-scale visual hierarchy of our model. Our Hierarchical Spatial Alignment (HSA) loss explicitly grounds micro-vision (what the robot feels) within local and macro visual contexts (what the robot sees).} %The three-scale visual hierarchy of our model. Our framework fuses information from three distinct visual modalities. Our Hierarchical Spatial Alignment (HSA) loss is designed to explicitly ground the micro-vision (what the robot feels) within the local and macro visual contexts (what the robot sees).}
    \label{fig:three-scale}
\end{figure}

We apply an InfoNCE contrastive loss to align the pooled tactile representation with its corresponding visual-region representations. The loss for aligning the tactile view with the wrist view is:
\begin{equation}
\mathcal{L}_{\mathrm{HSA-W}}
=
- \log
\displaystyle \frac{
\exp\!\left(\displaystyle \frac{h_\tau \cdot h_w}{\kappa} \right)
}{
\exp\!\left(\displaystyle \frac{h_\tau \cdot h_w}{\kappa} \right)
+
\displaystyle \sum_{i=1}^{N_k}
\exp\!\left(\displaystyle \frac{h_\tau \cdot h_{w,i}^{\mathrm{neg}}}{\kappa} \right)
},
\label{eq:hsa_w_loss}
\end{equation}
% \begin{equation}
% \mathcal{L}_{\text{HSA-W}}
% =
% - \log
% \frac{
% \exp\!\left( \frac{h_\tau \cdot h_w}{\kappa} \right)
% }{
% \exp\!\left( \frac{h_\tau \cdot h_w}{\kappa} \right)
% +
% \sum_{i=1}^{N_k}
% \exp\!\left( \frac{h_\tau \cdot h_{w,i}^{\text{neg}}}{\kappa} \right)
% }.
% \label{eq:hsa_w_loss}
% \end{equation}
where $h_{w,i}^{\mathrm{neg}}$ are $N_k$ negative samples (e.g., tokens from other regions or other images in the batch) and $\kappa$ is a temperature parameter. A similar loss, $\mathcal{L}_{\mathrm{HSA-TP}}$, is computed between $h_\tau$ and $h_{tp}$. The total alignment loss is:
\begin{equation}
\mathcal{L}_{\mathrm{HSA}}
=
\mathcal{L}_{\mathrm{HSA-W}}
+
\mathcal{L}_{\mathrm{HSA-TP}} .
\label{eq:hsa_total}
\end{equation}
This loss explicitly encourages the model to learn spatial correspondences between \textit{micro-vision} tactile images and localized regions in \textit{local} and \textit{macro-vision} camera views.

\subsubsection{Action Loss} The action expert is trained with a behavior cloning objective. Given the aligned multimodal tokens, it predicts an $H$-step action sequence
$\hat{A}^{(t)}$, which we supervise using the expert actions $A^{(t)}$.
We apply an $\ell_1$ loss over the horizon:
\begin{equation}
\mathcal{L}_{\mathrm{action}}
=
\frac{1}{H}
\sum_{j=0}^{H-1}
\left\lVert
\hat{a}^{(t)}_j - a^{(t)}_j
\right\rVert_1 .
\label{eq:action_loss}
\end{equation}
The total loss for Stage 1 is a weighted sum of these two objectives:
\begin{equation}
\mathcal{L}_{\mathrm{Stage 1}}
=
\mathcal{L}_{\mathrm{action}}
+
\lambda_{\mathrm{HSA}} \, \mathcal{L}_{\mathrm{HSA}} .
\label{eq:stage1_loss}
\end{equation}
This stage generates an initial policy trained on spatially aligned multimodal observations. Since future tactile predictions are not used during Stage 1, we provide a zero-valued embedding $H_\mathrm{null}$ in place of $H_\mathrm{dream}^{(t+N)}$, which is used in Stage 2.
% Upon completion of this stage, we have a competent initial policy that understands where its tactile sensor is and how to perform basic actions. Since no tactile prediction exists yet, we feed the policy a zero-tensor $H_\mathrm{null}$ in place of the input $H_\mathrm{dream}^{(t+N)}$.

\subsubsection{Pretrained Tactile World Model ($W_{\phi}$)}
A key component of our architecture is a pre-trained and frozen world model, $W_{\phi}$, which serves as a tactile feature extractor. Built upon V-JEPA 2~\cite{assran2025v}, the model is pretrained on a large-scale, unlabeled dataset of tactile image sequences, using self-supervised learning, to capture the underlying physical properties and dynamics of interactions. Given a tactile image $I_{\tau}$, the pretrained model extracts a latent representation $z_{\tau} = W_{\phi}(I_{\tau})$. 
% Given a tactile image $I_{\tau}$, $W_{\phi}$ encodes it into a latent representation $z_{\tau}$: 
% \begin{equation}
% z_{\tau}
% =
% W_{\phi}\!\left( I_{\tau} \right) .
% \label{eq:latent_encoder}
% \end{equation}

% Fig.~\ref{fig:heatmap} in the Appendix visualizes the evolution of the learned representations during pretraining. 

The tactile world model remains frozen throughout all subsequent training stages, providing a stable tactile representation space for action-conditioned latent forecasting in Stage 2. Additional pretraining details and representation visualizations are provided in Appendix~\ref{appendix:world_model} and Fig.~\ref{fig:heatmap}, respectively.

% We pre-train this model on a large, unlabeled dataset of tactile image sequences. $W_{\phi}$ (V-JEPA2~\cite{assran2025v}) is trained to be an expert in tactile physics. Its job is to take a tactile image $I_{\tau}$ and encode it in a rich, latent embedding $z_{\tau}$ that captures the underlying physical state (see Fig.~\ref{fig:heatmap} in Appendix for a visualization of the learned embedding across training).
% \begin{equation}
% z_{\tau}
% =
% W_{\phi}\!\left( I_{\tau} \right) .
% \label{eq:latent_encoder}
% \end{equation}
% Throughout all subsequent training stages, $W_{\phi}$ remains frozen, providing a stable and high-quality embedding of tactile information.

\subsection{Stage 2: Action Refining with a Dreamed Tactile Latent}

Building on Stage 1, we finetune the pretrained encoders ($E_{\psi}$) and policy ($\pi_{\theta}$) to refine actions based on anticipated contact interactions. %enable robust reasoning about physical interactions. 
We introduce a lightweight forecasting MLP ($F_{\eta}$) that predicts the latent tactile consequences of a draft action, while keeping the pretrained tactile world model ($W_{\phi}$) frozen as a stable feature extractor. % learns to dream of the latent sensory consequences of a draft action. 

Action refinement follows a three-step Think--Dream--Act framework:
\begin{itemize} [leftmargin=2em, itemsep=0.08em, topsep=0.15em]
    \item \textbf{THINK}: The policy $\pi_{\theta}$ generates a draft action $a_{\mathrm{draft}}^{(t)}$ based on the current aligned state $H_{\mathrm{align}}^{(t)}$ and a null tactile prediction $H_{\mathrm{null}}$.
    \item \textbf{DREAM}: The forecasting MLP $F_{\eta}$ predicts the future tactile embedding conditioned on the current tactile embedding $z_{\tau}^{(t)} = W_{\phi}(I_{\tau}^{(t)})$ and the draft action:
    \begin{equation}
        H_{\mathrm{dream}}^{(t+N)}
        = F_{\eta}\left(z_{\tau}^{(t)}, a_{\mathrm{draft}}^{(t)}\right).
    \end{equation}
    % , $H_{\mathrm{dream}}^{(t+N)}$. It takes two inputs: the current tactile embedding $z_{\tau}^{(t)}$ from the frozen $W_{\phi}$ and the draft action $a_{\mathrm{draft}}^{(t)}$ generated by the policy:
    % \begin{equation}
    % H_{\text{dream}}^{(t+N)}
    % =
    % F_{\eta}\!\left(
    % z_{\tau}^{(t)},\, a_{\text{draft}}^{(t)}
    % \right) .
    % \label{eq:dream_dynamics}
    % \end{equation}
     \item \textbf{ACT}: The predicted tactile embedding  $H_{\mathrm{dream}}^{(t+N)}$ is fed back into the policy $\pi_{\theta}$ along with the current aligned state %$H_{\mathrm{align}}^{(t)}$ 
     to produce the refined, final action $a_{\mathrm{final}}^{(t)}$.
\end{itemize}

During training, the forecasting MLP is supervised using future tactile embeddings extracted by the frozen tactile world model. The latent forecasting loss is defined as
\begin{equation}
    \mathcal{L}_{W}
    = \left\|
        H_{\mathrm{dream}}^{(t+N)}
        - \operatorname{sg}\left(W_{\phi}(I_{\tau}^{(t+N)})\right)
    \right\|_2^2,
\end{equation}
where $\operatorname{sg}(\cdot)$ denotes stop-gradient. We jointly optimize the action loss, Hierarchical Spatial Alignment loss, and latent forecasting loss:
\begin{equation}
    \mathcal{L}_{\mathrm{Stage\,2}}
    = \lambda_{\mathrm{action}}\mathcal{L}_{\mathrm{action}}
    + \lambda_{\mathrm{HSA}}\mathcal{L}_{\mathrm{HSA}}
    + \lambda_W\mathcal{L}_W.
\end{equation}
The draft action is detached during forecasting, while $E_{\psi}$, $\pi_{\theta}$, and $F_{\eta}$ are jointly optimized, with $W_{\phi}$ remaining frozen. Details can be found in Algorithm~\ref{alg:dreamtacvla} in Appendix.

%% file: sec/4_experiments.tex
\section{Experiments}
\label{sec:experiments}
%More about simulation setup
%Success rate table in simulation?
%Simulation tasks of tactile images

We hypothesize that robust contact-rich manipulation requires an agent to not only react to physical interactions but also anticipate their consequences. This capability is enabled by combining (1) spatially grounded perception of current observations through Hierarchical Spatial Alignment (HSA) and (2) a predictive world model that anticipates future tactile outcomes. Our experiments evaluate the contributions of both components.
% Our evaluation seeks to answer three primary research questions, each building upon the last:
% \begin{enumerate}[label=\arabic*), leftmargin=2em]
%     \item Does our full model, DreamTacVLA, outperform state-of-the-art VLAs (which only contains vision and depth \cite{qu2025spatialvla} \cite{zhang2025dreamvla}) on complex, contact-rich tasks?
%     \item Is the Hierarchical Spatial Alignment (HSA) component necessary? How does it compare to a model trained without this explicit cross-modal grounding?
%     \item Does the think-dream-act pipeline, enabled by the world model, provide a measurable benefit over a policy that only reacts to the present (our Stage 1 model)?
% \end{enumerate}

\subsection{Experimental Setup}
\label{sec:exp_setup}

\textbf{Implementation Details.} 
The full system consists of a CLIP-based language-conditioned policy~\cite{pmlr-v139-radford21a}, modality-specific encoders, a frozen V-JEPA 2-based tactile world model~\cite{assran2025v} with lightweight residual adapters, and a transformer-based action expert that predicts 7-DOF actions over a 45-step horizon. The adapter adds only 5.5M trainable parameters (1.8\% overhead) to the frozen ViT-L. Full architecture, adapter, and optimization details are provided in Appendix~\ref{appendix:implementation_details}.

% \vspace{1mm}
% \noindent\textbf{III. Training Strategy}

% \noindent We apply a two-stage training process for each batch of data.

% \textbf{Stage 1:} The ResNet18 camera backbones are activated, and our HSA ($H_{align}^{(t)}$) is computed. However, the world model is deactivated at this stage; we substitute its output with an all-zero tensor, $H_{null}(\text{Zeros})$. At the end of Stage 1, the action expert infers an initial draft action, $a_{draft}^{(t)}$. CLIP only learns from wrist-tactile loss $H_{}$

% \textbf{Stage 2:} The world model is activated to process the tactile image. Its output is then used as an auxiliary input, combined with the draft action $a_{draft}^{(t)}$ from Stage 1, and fed into the final MLP. HSA is also computed in this stage,

% \textbf{Optimizers and Loss:} The ResNet backbones and policy are trained using the AdamW optimizer with a batch size of 16. We use a cosine learning rate decay schedule, starting with an initial learning rate of $1 \times 10^{-4}$ for the policy and $1 \times 10^{-5}$ for the encoders. The combined loss weights are set to $\lambda_{HSA}=0.1$ and $\lambda_{W}=1.0$.

\textbf{Simulation \& Hardware.} We conduct experiments in both simulation and the real world. Our simulation environment is built in IsaacSim~\cite{NVIDIA_Isaac_Sim} with a physics-based tactile sensor model based on TacEx~\cite{nguyen2024tacex}, following the Taxim's optical, texture-based approach that synthesizes gel deformation via light-transport modeling and marker-texture warping~\cite{9681378}. This yields realistic, high-resolution tactile images closely mimicking the real sensor data, enabling large-scale data collection (1000 demonstrations per task) with randomized object poses in parallel environments. Our real-world setup uses a Dobot XTrainer platform with a parallel gripper, one DIGIT sensors~\cite{lambeta2020digit}, and two Realsense D405 cameras as wrist and third-person cameras, with 100 expert demonstrations per real-world task. All experiments use a single arm with one gripper-mounted DIGIT tactile sensor throughout. Figure~\ref{fig:predictions} in the Appendix provides a qualitative comparison of simulation and real-world data streams. We run the inference loop at 50 Hz.

% \begin{figure}
%     \centering
%     \includegraphics[width=0.98\linewidth]{CVPR2026/figures/xtrainer.jpg}
%     \caption{Experiment platform in IsaacSim and realworld}
%     \label{fig:xtrainer}
% \end{figure}

% \noindent\textbf{Tasks.} We evaluate on four challenging contact-rich tasks.
% \begin{itemize}[label={}]
%     \item Peg-in-Hole: A classic robotics task requiring high precision. The port is partially occluded, forcing the policy to rely on tactile feedback for final alignmensinglet.
%     \item \textbf{USB Insertion:} Inserting a USB-A plug into a port. This task has extremely tight tolerances (sub-millimeter) that are ambiguous from vision alone.
%     \item \textbf{Gear Assembly:} Sliding a small gear onto a shaft. This requires aligning the gear's hole with the shaft, a task easily failed by misalignment.
%     \item \textbf{Pen Stabilization:} The agent must grasp a pen and keep its tip stable at a target 3D coordinate, resisting small perturbations. Success is measured by minimizing the end-effector deviation from the target.
% \end{itemize}

\textbf{Tasks.} We evaluate four challenging contact-rich manipulation tasks, illustrated in Figure~\ref{fig:sim_real1} in the Appendix. 
1) Peg-in-Hole: Pick, place, insert a cuboid into a hole with 1 mm clearance. 
The hole is partially occluded during the placement and insertion, requiring tactile feedback for precise alignment. 
2) USB Insertion: Pick, place, insert a USB-A plug into a port with submillimeter tolerances, where visual observations alone are insufficient for precise alignment.
3) Gear Assembly: Pick, place, and insert a gear onto a shaft, requiring precise alignment between the gear's central hole and the shaft. 
4) Tool Stabilization: Grasp a cube and use one of its vertices to support a thin vertical cylinder on a tabletop, maintaining its upright pose under small disturbances. All four tasks begin with a pick-and-place phase before the contact-rich insertion or stabilization stage. Among these tasks, we construct a hybrid dataset comprising approximately 80\% simulated and 20\% real-world demonstrations (see Fig.~\ref{fig:data-composition} in the Appendix for the detailed composition). These simulated demonstrations are used solely to pre-train the tactile world model $W_\phi$; the action expert $\pi_\theta$ is trained only on the real-world demonstrations. This keeps the policy grounded in real sensor statistics while allowing the frozen tactile world model benefit from large-scale simulated contact data.

\textbf{Baselines.}
We compare DreamTacVLA against state-of-the-art policies and controlled ablations of our own method. External baselines include ACT~\cite{Zhao-RSS-23}, Diffusion Policy~\cite{chi2025diffusion} and $\pi0$~\cite{black2024pi_0}. We also include a standard ``ACT + Tactile'' baseline to isolate whether the performance gain comes from the ``Dreaming'' architecture or simply from the availability of tactile data.
We also evaluate several variants of our model:
\begin{itemize} [leftmargin=2em, itemsep=0.08em, topsep=0.15em]
\item Ours (HSA-Only, No Dream): 
Uses HSA-aligned encoders relying only on the current states, removing the contribution of the tactile world model.
\item Ours (No HSA, Dream-Only): Ablation trained without the $\mathcal{L}_{\mathrm{HSA}}$ loss, used to test whether spatial alignment can be learned implicitly.
\item Ours (HSA \& Dream): Our full two-stage model including both HSA and Think–Dream–Act.% pipeline.
\end{itemize}

% We compare DreamTacVLA against strong state-of-the-art policies and controlled ablations of our own method. External baselines include ACT~\cite{Zhao-RSS-23}, Diffusion Policy~\cite{chi2025diffusion} and $\pi0$~\cite{black2024pi_0}. We also insclude a standard ``ACT + Tactile'' baseline to isolate whether the performance gain comes from the complex ``Dreaming'' architecture or simply from the availability of tactile data.
% We also evaluate several variants of our model:\\
% Ours (HSA-Only, No Dream): Stage-1 variant that uses HSA-aligned encoders but relies solely on the current state, removing the contribution of the world model.\\
% Ours (No HSA, Dream-Only): Ablation trained without the $\mathcal{L}_\mathrm{HSA}$ loss, used to test whether spatial alignment can be learned implicitly.\\
% Ours (HSA \& Dream): Our full Stage-2 model incorporating both the HSA and the Think–Dream–Act pipeline.

\subsection{Main Results}
\label{sec:main_results}

\begin{table*}[tb]
    \centering
    \caption{Task success rates (\%) in real-world (100 trials per run). Results are reported as mean $\pm$ standard deviation over 3 runs.}
    \label{tab:main_results}
    \resizebox{\textwidth}{!}{
    \begin{tabular}{@{}lcccc@{}}
        \toprule
        \textbf{Model} & \textbf{Peg-in-Hole} & \textbf{USB Insert} & \textbf{Gear Assembly} & \textbf{Tool Stabilize} \\
        \midrule
        ACT \cite{Zhao-RSS-23} & $35.2 \pm 0.7\%$ & $62.6 \pm 0.5\%$ & $22.4 \pm 0.8\%$ & $19.3 \pm 0.6\%$ \\
        Diffusion Policy \cite{chi2025diffusion} & $35.5 \pm 0.9\%$ & $56.3 \pm 0.8\%$ & $33.1 \pm 0.7\%$ & $30.4 \pm 0.9\%$ \\
        $\pi 0$ \cite{black2024pi_0} & $48.7 \pm 1.0\%$ & $59.4 \pm 0.9\%$ & $45.2 \pm 1.1\%$ & $41.0 \pm 0.8\%$ \\
        ACT (w/ Tactile) & $45.6 \pm 0.3\%$ & $63.1 \pm 0.4\%$ & $40.2 \pm 0.2\%$ & $39.1 \pm 0.7\%$ \\
        \midrule
        Ours (HSA-Only, No Dream) & $60.8 \pm 0.9\%$ & $63.7 \pm 0.8\%$ & $51.5 \pm 1.0\%$ & $42.9 \pm 0.7\%$ \\
        Ours (No HSA, Dream-Only) & $75.4 \pm 0.8\%$ & $75.2 \pm 0.7\%$ & $64.9 \pm 0.6\%$ & $68.5 \pm 0.9\%$ \\
        \textbf{Ours (HSA \& Dream)} & $\mathbf{95.0 \pm 0.2\%}$ & $\mathbf{85.7 \pm 0.6\%}$ & $\mathbf{81.1 \pm 0.4\%}$ & $\mathbf{74.6 \pm 0.5\%}$ \\
        \bottomrule
    \end{tabular}
    }
\end{table*}
\textbf{Real-World Results.} Table~\ref{tab:main_results} presents the success rates across four real-world contact-rich manipulation tasks. DreamTacVLA consistently outperforms all baselines, achieving success rates of 95.0\%, 85.7\%, 81.1\%, and 74.6\% on Peg-in-Hole, USB Insertion, Gear Assembly, and Tool Stabilization, respectively, averaging 84.1\% and outperforming the strongest baseline, $\pi_0$ (48.6\%), by 35.5\%. The largest improvement occurs in Peg-in-Hole (+46.3\%), where partial occlusion and tight tolerances make visual alignment particularly challenging. While incorporating tactile sensing improves ACT's average success rate from 34.9\% to 47.0\%, its performance remains substantially below ours, suggesting that effective contact-rich manipulation requires more than simply adding tactile observations. Moreover, the relative performance of our ablated variants is consistent with simulation results in Appendix~\ref{app:sim_results}, providing further evidence of the benefits of spatial alignment and tactile prediction across simulated and physical environments.

% Vision-only baselines (ACT~\cite{Zhao-RSS-23}, Diffusion Policy~\cite{chi2025diffusion}) perform poorly, especially on tasks like USB Insertion and Gear Assembly, where visual ambiguity and depth occlusion are significant. The Diffusion Policy baseline demonstrates moderate competence but fails to capture the fine-grained temporal consistency required for stable contact handling, often oscillating or prematurely retracting during insertion. The ACT with tactile baseline shows that tactile modality can improve performance; by comparison, our HSA and ``Dream'' method make more effective use of tactile feedback.

% DreamTacVLA consistently outperforms all baseline and ablated models, with the best performance observed in USB Insertion and Peg-in-Hole. Both tasks demand precise micro-slip perception, iterative pose refinement, and stable contact maintenance. These capabilities are difficult to achieve with vision-only or feedforward tactile policies. The Think-Dream-Act method is particularly influential in these settings: as the end-effector approaches the socket or hole, the policy executes controlled residual adjustments rather than committing to a simple open-loop motion. This behavior indicates that the tactile world model provides %high-frequency
% predictive feedback that guides fine-grained corrections.

\textbf{Effect of HSA and World Model.} The ablation results in Table~\ref{tab:main_results} demonstrate the complementary contributions of spatial alignment and tactile prediction. HSA alone achieves an average success rate of 54.7\%, while Dream-only improves this to 71.0\%, highlighting the benefits of anticipating future contact states beyond relying on current observations. Combining both components further increases success to 84.1\%, outperforming HSA-only and Dream-only by 29.4 and 13.1 percentage points, respectively. Similar trends are observed in simulation, where the full model achieves 96.6\%, compared with 86.7\% and 91.9\% for the respective ablations. These findings suggest that spatially grounding tactile feedback within visual observations and predicting its action-conditioned tactile evolution jointly improve contact-rich manipulation.

\subsection{Additional Experiments}
\label{sec:ablations}

%World model size effect
%Tactile data size effect

% We conduct detailed ablations to validate our key design choices.

\begin{figure}[tb]
    \centering
    \includegraphics[width=0.8\linewidth]{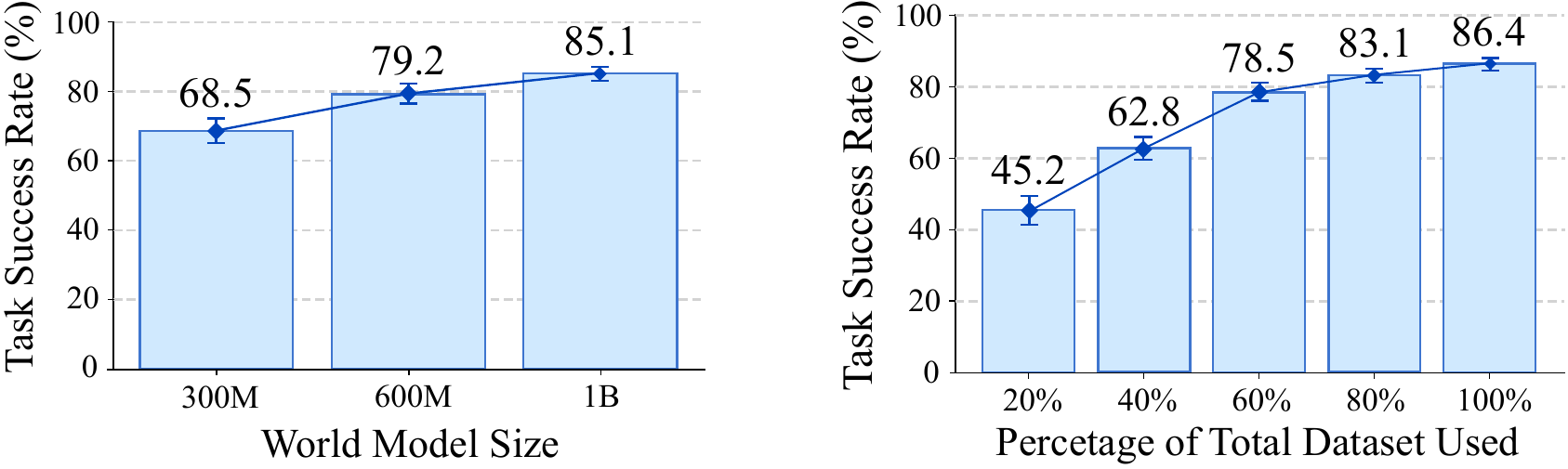}
    \caption{Evaluation on the effects of world model size and data scaling.}
    \label{fig:ablation_study}
    % \vspace{-1em}
\end{figure}

\noindent\textbf{World Model Sizes.} 
% We ablate the components of our world model ($\mathcal{L}_W$). As shown in Figure \ref{fig:placeholder}, training a world model to predict \textit{only} future visual images (like DreamVLA \cite{NEURIPS2025_22d4f952}) provides a minor boost. The tactile forecast is the most critical component. However, training the model to predict \textit{all} future modalities (Tactile+Vision) yields the best results, as it learns a more consistent cross-modal physics model.
We scale the world model $W_{\phi}$ from 300M to 1B parameters while keeping the architecture and the training data fixed. As shown in the left panel of Figure \ref{fig:ablation_study}, task success rate improves consistently with model capacity, from 68.5\% at 300M to 79.2\% at 600M and 85.1\% at 1B, with the largest gain ($+10.7\%$) from 300M to 600M. %and diminishing yet positive returns thereafter. 
We hypothesize that larger models provide more accurate latent tactile prediction, enabling more effective action refinement during the dreaming stage. These results suggest that further scaling may benefit contact-rich manipulation. %We attribute this to higher-capacity models producing more accurate latent tactile forecasts, which supply a stronger signal for action refinement during the dreaming stage. The sustained gains suggest that further scaling could continue to benefit contact-rich manipulation.

% \begin{table}[h]
%     \centering
%     \caption{Ablation on World Model Size, it shows the power of the world model.}
%     \label{tab:ablation_dream}
%     \resizebox{\columnwidth}{!}{
%     \begin{tabular}{@{}lcc@{}}
%         \toprule
%         \textbf{World Model Size} & \textbf{USB Insert (SR\%)} & \textbf{Pen Stabilize (SR\%)} \\
%         \midrule
%         300M & 61\% & 68\% \\
%         600M & 65\% & 70\% \\
%         1B & 78\% & 81\% \\
%         \bottomrule
%     \end{tabular}
%     }
% \end{table}

\noindent\textbf{%Tactile 
Data Amount.}
We train separate model instances on progressively larger subsets (20\%--100\%) of the collected data. As shown in the right panel of Figure~\ref{fig:ablation_study}, success rate improves with increasing data and begin to plateau around 60\%, suggesting that our dataset provides sufficient coverage for these tasks, while further data scaling may yield modest gains. %consistently with more data and begins to converge around 60\% of the dataset, indicating our collection is sufficient for these tasks while the continued slight upward trend suggests further scaling could yield diminishing marginal gains.

%% file: sec/5_conclusion.tex
\section{Limitations}
The Think--Dream--Act pipeline adds inference latency, since it requires a draft policy pass followed by a world-model rollout before the final action, potentially limiting its applicability to high-frequency control. Moreover, our evaluation is limited to four contact-rich tasks using a DIGIT tactile sensor, leaving generalization to other sensors and broader task distributions unexplored. Finally, the world model remains frozen after pretraining, bounding its adaptation to unseen contact interactions.

\section{Conclusion}

We present DreamTacVLA, a physically grounded Vision-Language-Action framework that addresses the contact-blindness of vision-centric policies. The method combines a Hierarchical Spatial Alignment (HSA) loss that tightly grounds tactile, wrist, and third-person cues, and a Think–Dream–Act strategy that uses a tactile world model to forecast tactile outcomes of draft actions, enabling anticipatory contact reasoning.

Across four contact-rich manipulation tasks, DreamTacVLA outperforms vision-only and vanilla vision + tactile baselines, achieving near-perfect performance in simulation and real world. Ablations confirm the complementary benefits of tactile grounding and tactile forecasting.

Although Think–Dream–Act introduces inference overhead, future work will explore policy distillation and adaptive dreaming for faster single-pass reasoning. Scaling tactile world models with larger multimodal corpora offers a path toward more general agents that can reason about physical interactions with human-like intuition.

%% file: sec/6_supplementary.tex
\clearpage
\section*{Appendix Catalog}
\addcontentsline{toc}{section}{Appendix Catalog}

This appendix provides additional implementation details, experimental results, and theoretical background for DreamTacVLA. Below is a catalog of the contents:

     \textbf{Appendix A: Simulation Pipeline and Large-Scale Data Collection} \dotfill \pageref{appendix:sim} \\ 
     \textbf{Appendix B: Implementation Details} \dotfill \pageref{appendix:implementation_details}\\
     \textbf{Appendix C: Hierarchical Spatial Alignment (HSA)} \dotfill \pageref{appendix:HSA}\\
     \textbf{Appendix D: Tactile World Model Pretraining and Forecasting}  \dotfill \pageref{appendix:world_model} \\
     \textbf{Appendix E: Evaluation Protocol} \dotfill \pageref{appendix:evaluation}\\
     \textbf{Appendix F: Extended Related Work} \dotfill \pageref{appendix:extended_related_work}

\section{Simulation Pipeline and Large-Scale Data Collection}\label{appendix:sim}

This section clarifies the role of simulation in DreamTacVLA and explains
data quality, expert reliability, and sim-to-real validity.
Simulation is used not as a proxy for deployment, but as a scalable and stable substrate
for learning spatially grounded tactile representations and tactile future prediction.

\subsection{IsaacSim Environment}

All simulation experiments are conducted in NVIDIA IsaacSim with GPU-accelerated PhysX,
using conservative physics settings to ensure stable long-horizon contact under tight tolerances.
The simulated robot is a digital twin of the Dobot XTrainer with shared camera extrinsics between
simulation and real-world setups.
Following the main submission, we employ a high-fidelity visuotactile simulator based on
TacEx and Taxim-style optical and marker-texture warping to synthesize
GelSight-like tactile observations, enabling large-scale data collection under randomized object
poses in parallel simulation environments.

\begin{table}[h]
\centering
% \small
\caption{Simulation configuration (IsaacSim).}
\label{tab:sim_config}
\begin{tabular}{l l}
\toprule
\textbf{Component} & \textbf{Setting} \\
\midrule
Physics timestep & $1/60$ s \\
Solver substeps & 2 \\
Control frequency & 30 Hz \\
Max episode length & 300 steps \\
Parallel environments & 1024 / GPU(24G) \\
Sensors & TPV RGB, Wrist RGB, Tactile \\
\bottomrule
\end{tabular}
\end{table}

\subsection{Automated Expert Demonstrations (cuRobo)}

Expert demonstrations are not collected via human teleoperation.
Instead, we employ an automated, privileged expert based on cuRobo, which has
access to ground-truth object pose, collision geometry, and task-specific termination conditions.
The expert operates in task space and generates collision-aware trajectories that satisfy
geometric alignment and insertion constraints.
These trajectories are time-parameterized (minimum-jerk) and resampled at the control frequency
(30\,Hz) to produce per-step 6D end-effector relative pose + gripper supervision,
consistent with the action representation used by the policy, resulting in
low-noise and consistent demonstrations for contact-rich manipulation.

\begin{table}[h]
\centering
% \small
\caption{cuRobo expert trajectory generation.}
\label{tab:curobo}
\begin{tabular}{l l}
% \begin{tabular}{p{0.35\linewidth} p{0.55\linewidth}}
\toprule
\textbf{Aspect} & \textbf{Specification} \\
\midrule
Planner & cuRobo (batched) \\
Batch size & 256 queries \\
Planning horizon & 32 waypoints (minimum-jerk) \\
Collision checking & Robot + objects + table \\
IK tolerance & 2 mm position / $2^\circ$ rotation \\
Action format & 6D EE delta pose + gripper \\
\bottomrule
\end{tabular}
\end{table}

This design guarantees demonstration quality:
stable contact behavior at sub-millimeter precision would be extremely difficult to achieve
via human teleoperation without force or tactile feedback.

\subsection{Success-Based Filtering and Synchronization}

Only successful rollouts are retained as demonstrations, based on task-specific
geometric and temporal criteria (e.g., insertion depth, alignment error, sustained stability).
All modalities are logged with strict step-level synchronization; if any modality is missing
at a timestep, the entire episode is discarded. This filtering is applied only for constructing expert supervision for behavior cloning;
evaluation is performed on fresh randomized initializations.

\subsection{Performance within IsaacSim}
\label{app:sim_results}
To assess the effectiveness of our method under controlled conditions, we report
task success rates measured inside IsaacSim using the same evaluation protocol
as in real-world experiments as Table \ref{tab:isaacsim_results} shows.
\begin{table*}[h]
\centering
\caption{Task success rate (\%) in IsaacSim (100 trials per task).}
\label{tab:isaacsim_results}
\begin{tabular}{l c c c c}
\toprule
\textbf{Model} & \textbf{Peg-in-Hole} & \textbf{USB Insert} & \textbf{Gear Assembly} & \textbf{Tool Stabilize} \\
\midrule
ACT & 72.4 & 78.1 & 65.3 & 61.7 \\
Diffusion Policy & 75.6 & 80.2 & 69.4 & 66.1 \\
$\pi_0$ & 83.5 & 85.9 & 77.2 & 74.6 \\
Ours (HSA-Only, No Dream) & 89.1 & 90.4 & 84.6 & 82.7 \\
Ours (No HSA, Dream-Only) & 93.8 & 94.6 & 88.9 & 90.1 \\
\textbf{Ours (HSA \& Dream)} & \textbf{98.6} & \textbf{97.9} & \textbf{95.2} & \textbf{94.7} \\
\bottomrule
\end{tabular}
\end{table*}
These simulation-side results provide additional context that the relative performance trends
and ablation ordering observed on real hardware are already consistent in IsaacSim.
This consistency suggests that the benefits of HSA and Dreaming are not solely driven by
real-world noise or dataset imbalance.

\begin{algorithm}[t]
\caption{DreamTacVLA: Two-Stage Training with Think--Dream--Act} %(Implementation-Matched)}
\label{alg:dreamtacvla}
\begin{algorithmic}[1]
\Require Dataset $\mathcal{D}_1$ consists of tuples $(L, s^{(t)}, I^{(t)}_{w}, I^{(t)}_{tp}, I^{(t)}_{\tau}, a_t, B^{(t)})$
\Require Dataset $\mathcal{D}_2$ consists of tuples $(L, s^{(t)}, I^{(t)}_{w}, I^{(t)}_{tp}, I^{(t)}_{\tau}, a_t, I^{(t+N)}_{\tau}, B^{(t)})$
\Require Modules: multimodal encoders $E_\psi$, policy $\pi_\theta$, tactile world encoder $W_\phi$ (pretrained, frozen), forecasting MLP $F_\eta$
\Require Hyperparams: $S_1,S_2,\lambda_{\mathrm{HSA}},\lambda_W,\lambda_{\mathrm{action}}$; null-dream token $H_{\mathrm{null}}$

\State Initialize parameters of $E_\psi$ and $\pi_\theta$
\State Set $H_{\mathrm{null}}$ (zeros or learned)
\Statex

\State \textbf{Stage 1 (no dreaming): pre-train encoders + policy with Action + HSA}
\For{$s \leftarrow 1$ to $S_1$}
    \State Sample $(\mathcal{L}, s^{(t)}, I^{(t)}_{w}, I^{(t)}_{tp}, I^{(t)}_{\tau}, a_t, B^{(t)})$ from $\mathcal{D}_1$
    \State $H^{(t)}_{\mathrm{align}} \leftarrow E_\psi(\mathcal{L}, I^{(t)}_{w}, I^{(t)}_{tp}, I^{(t)}_{\tau}, s^{(t)})$
    \State $\mathcal{L}_{\mathrm{HSA}} \leftarrow \textsc{InfoNCE}(H^{(t)}_{\mathrm{align}}, B^{(t)})$
    \State $\hat{a}_t \leftarrow \pi_\theta(H^{(t)}_{\mathrm{align}}, H_{\mathrm{null}})$
    \State $\mathcal{L}_{\mathrm{action}} \leftarrow \mathcal{L}_\textsc{Diffusion}(\hat{a}_t, a_t)$
    \State $\mathcal{L} \leftarrow \mathcal{L}_{\mathrm{action}} + \lambda_{\mathrm{HSA}} \mathcal{L}_{\mathrm{HSA}}$
    \State Backpropagate $\mathcal{L}$ and update $\psi,\theta$
\EndFor
\Statex

\State \textbf{Stage 2: finetune with latent dreaming (Think--Dream--Act)}
\State Load pretrained $W_\phi$ and freeze it (no gradient updates)
\State Initialize $F_\eta$ (trainable)
\For{$s \leftarrow 1$ to $S_2$}
    \State Sample $(\mathcal{L}, s^{(t)}, I^{(t)}_{w}, I^{(t)}_{tp}, I^{(t)}_{\tau}, a_t, I^{(t+N)}_{\tau}, B^{(t)})$ from $\mathcal{D}_2$
    \State $H^{(t)}_{\mathrm{align}} \leftarrow E_\psi(\mathcal{L}, I^{(t)}_{w}, I^{(t)}_{tp}, I^{(t)}_{\tau}, s^{(t)})$
    \State $\mathcal{L}_{\mathrm{HSA}} \leftarrow \textsc{InfoNCE}(H^{(t)}_{\mathrm{align}}, B^{(t)})$

    \State \textbf{THINK:} compute draft action (stop gradient)
    \State $a^{(t)}_{\mathrm{draft}} \leftarrow \textsc{StopGrad}\!\left(\pi_\theta(H^{(t)}_{\mathrm{align}}, H_{\mathrm{null}})\right)$

    \State \textbf{DREAM:} forecast tactile latent (no grad through $W_\phi$)
    \State $z^{(t)}_{\tau} \leftarrow W_\phi(I^{(t)}_{\tau})$
    \State $\tilde{z}^{(t+N)}_{\tau} \leftarrow F_\eta(z^{(t)}_{\tau}, a^{(t)}_{\mathrm{draft}})$
    \State $z^{(t+N)}_{\tau} \leftarrow \textsc{StopGrad}\!\left(W_\phi(I^{(t+N)}_{\tau})\right)$
    \State $\mathcal{L}_W \leftarrow \left\|\tilde{z}^{(t+N)}_{\tau} - z^{(t+N)}_{\tau}\right\|_2^2$

    \State \textbf{ACT:} condition policy on dreamed latent
    \State $a^{(t)}_{\mathrm{final}} \leftarrow \pi_\theta(H^{(t)}_{\mathrm{align}}, \tilde{z}^{(t+N)}_{\tau})$
    \State $\mathcal{L}_{\mathrm{action}} \leftarrow \mathcal{L}_\textsc{{Diffusion}}(a^{(t)}_{\mathrm{final}}, a_t)$

    \State $\mathcal{L} \leftarrow \lambda_{\mathrm{action}}\mathcal{L}_{\mathrm{action}} + \lambda_{\mathrm{HSA}}\mathcal{L}_{\mathrm{HSA}} + \lambda_W \mathcal{L}_W$
    \State Backpropagate $L$ and update $\psi,\theta,\eta$ (keep $\phi$ frozen; no grad through $a^{(t)}_{\mathrm{draft}}$)
\EndFor
\end{algorithmic}
\end{algorithm}

\section{Implementation Details}
\label{appendix:implementation_details}

\label{app:dataset-composition}
\begin{figure}[htbp]
    \centering
     \includegraphics[width=0.6\linewidth]{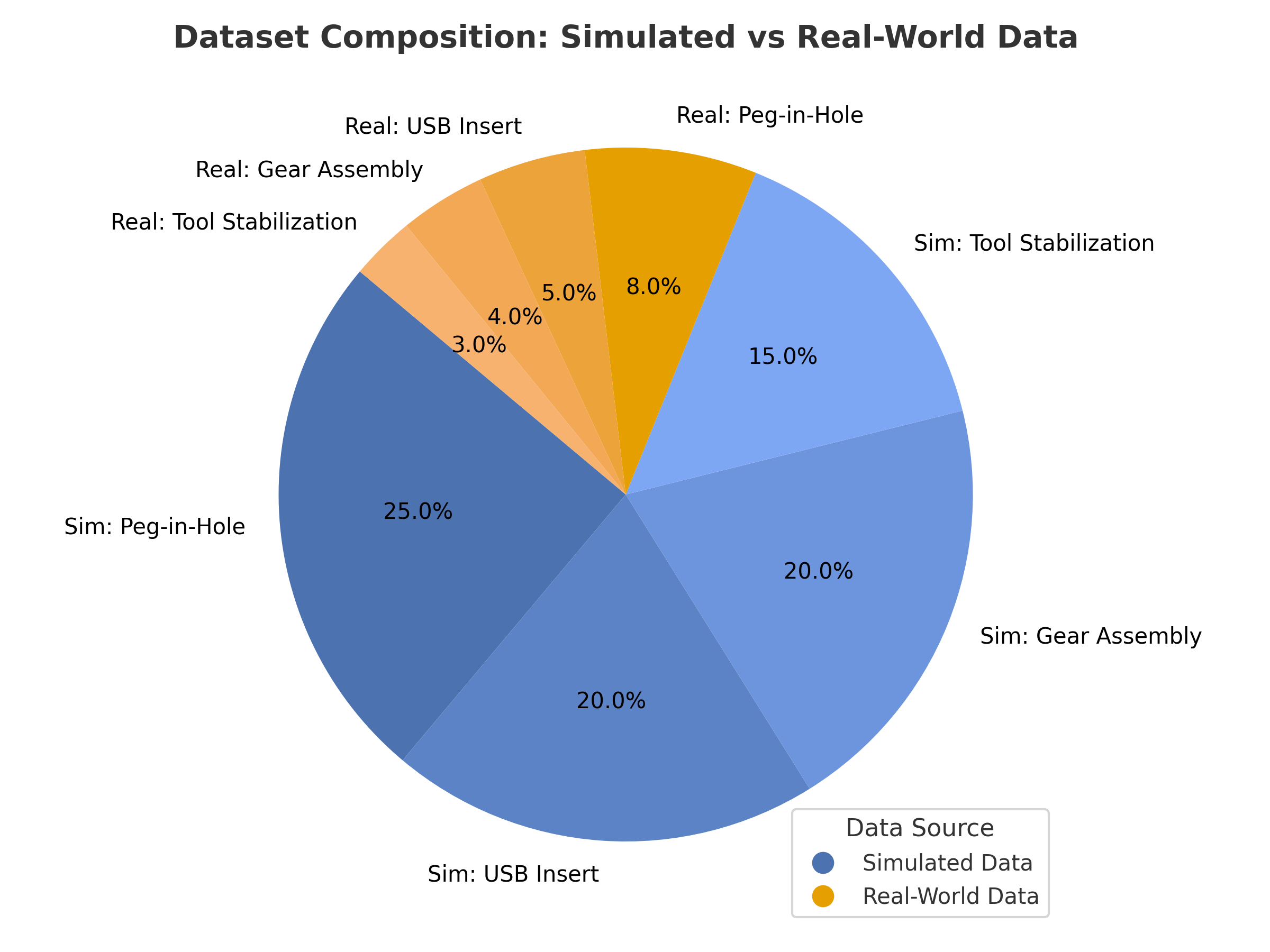}
    \caption{The dataset consists of 80\% simulated demonstrations and 20\% real-world demonstrations, each containing four task categories: Peg-in-Hole, USB Insert, Gear Assembly, and Tool Stabilization. Blue segments represent simulated data, while orange segments denote real-world data.}
    \label{fig:data-composition}
\end{figure}

\begin{figure}[htbp]
    \centering
    \includegraphics[width=0.8\linewidth]{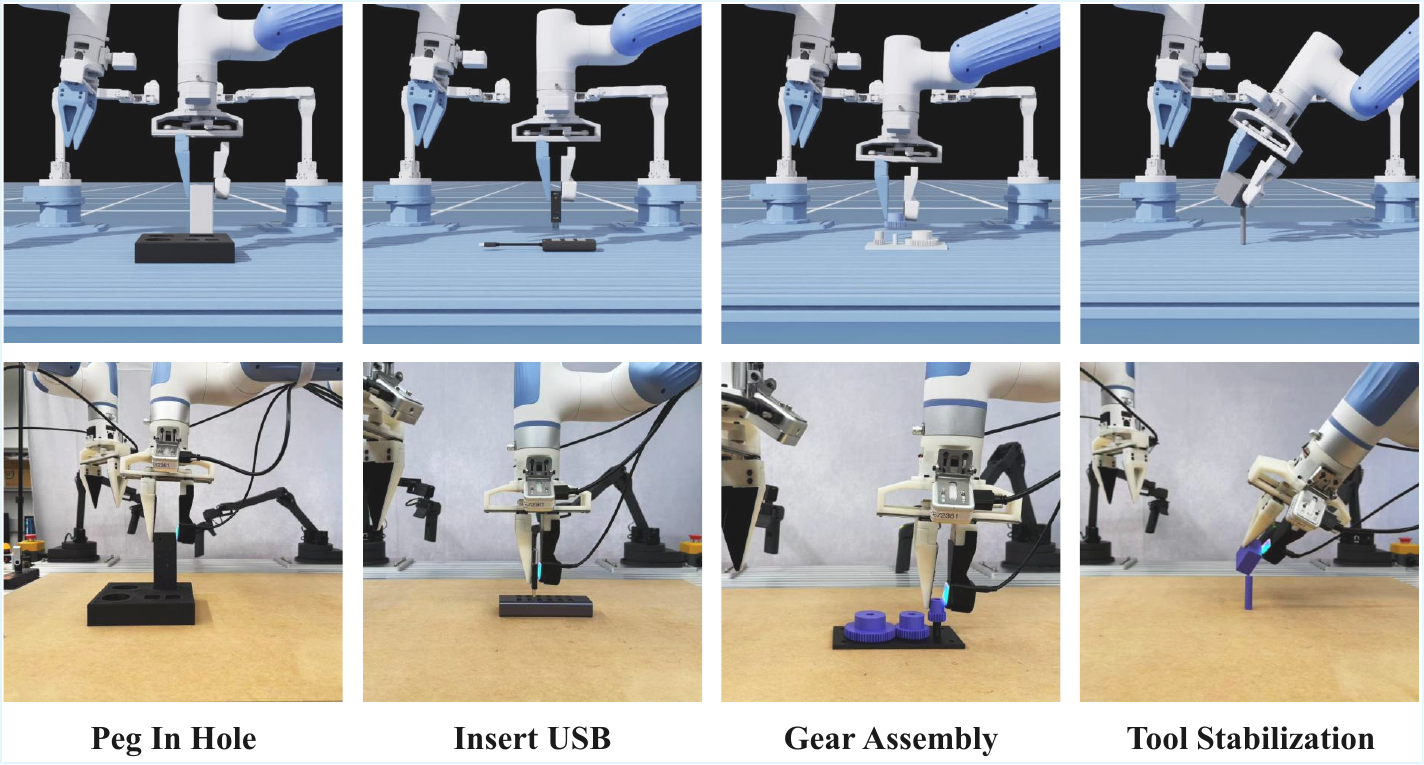}
    \caption{Tasks used to evaluate DreamTacVLA. From left to right: Peg-in-Hole, USB Insert, Gear Assembly, and Tool Stabilization. Each task demands precise, contact-rich manipulation, including aligning tight tolerances, detecting slip, or maintaining stable contact.}
    \label{fig:sim_real1}
\end{figure}

\begin{figure}[htbp]
    \centering
    \includegraphics[width=\linewidth]{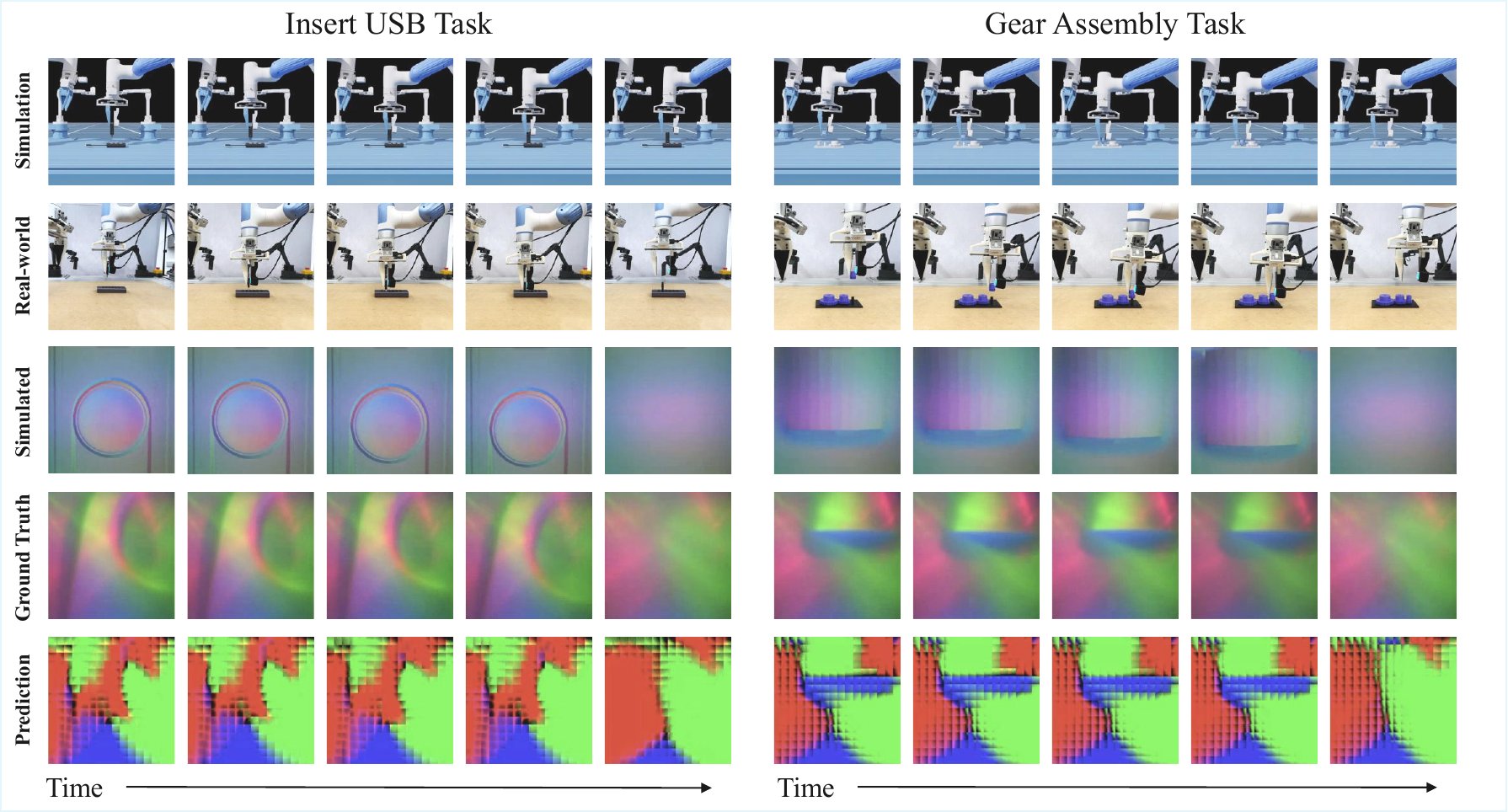}
    \caption{Qualitative comparison of our model's tactile prediction. For both the Peg-in-Hole and Tool Stabilization tasks, we visualize the sequence (left to right) comparing our model's Prediction (bottom row) to the Ground Truth tactile data (fourth row). The corresponding tactile images are provided as well.}
    \label{fig:predictions}
\end{figure}

\begin{figure}[htbp]
    \centering
    \includegraphics[width=0.7\linewidth]{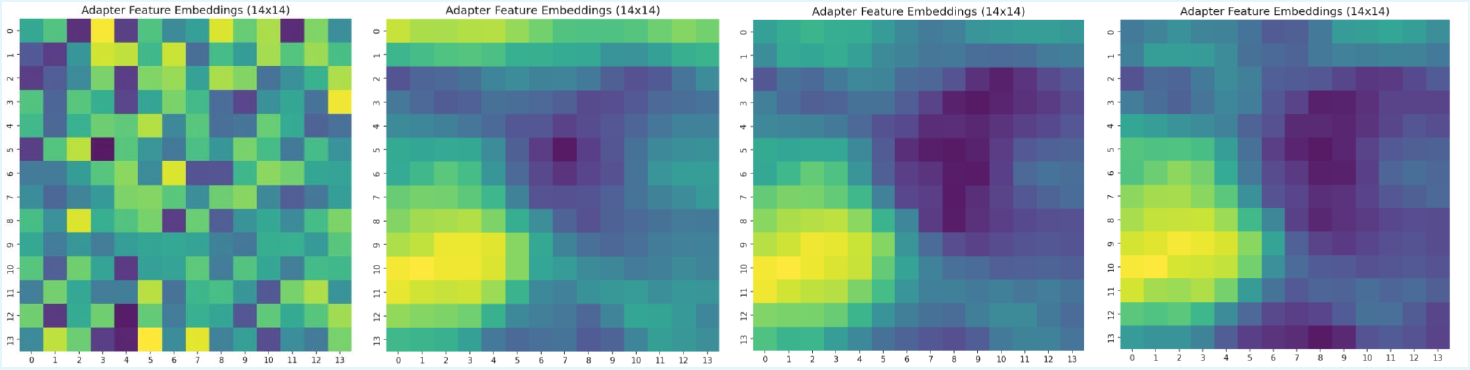}
    \caption{Visualization of the world model’s predicted future-state embedding $W_{\phi}(I_{\tau})$ across training. Initially, the embedding is noisy and unstructured, indicating weak predictive ability. As training advances, the embedding becomes increasingly concentrated and stable, revealing that the world model is learning a coherent representation of future tactile–visual dynamics.}
    \label{fig:heatmap}
\end{figure}

% \vspace{1mm}
% \noindent\textbf{Model Architecture.}
% Policy and Encoders: The policy $\pi_{\theta}$ (Language Backbone) is initialized from a pretrained CLIP (clip-vit-large-patch14) model \cite{pmlr-v139-radford21a} and finetuned on our dataset. This CLIP model is also responsible for aligning wrist camera and tactile images.  The tactile image ($I_{\tau}$) encoder is a V-JEPA2 model \cite{assran2025v} (ViTL/ViTG), initialized from its official pre-trained weights.

% \noindent Action Expert: Our action expert is an action transformer, which is trained to predict a 7-DOF action (6D end-effector pose + 1D gripper state) over a 45-step horizon. The same horizon is used during inference.

\subsection{Model Architecture}
\label{sec:architecture}

\paragraph{Overview.}
DreamTacVLA uses a hybrid vision-language-action architecture that combines Action Chunking with Transformers (ACT)~\cite{Zhao-RSS-23} as the policy backbone with V-JEPA 2~\cite{assran2025v} vision encoders for tactile prediction. The architecture processes multi-modal sensory inputs, including RGB images from third-person and wrist cameras, and tactile images from vision-based tactile sensors mounted on the gripper.

\paragraph{Visual Encoders.}
For RGB camera inputs, we use frozen ResNet-18~\cite{he2016deep} backbones. Each RGB camera (top view and wrist cameras) has a dedicated ResNet-18 encoder. The backbone outputs are projected to the transformer hidden dimension via a $1 \times 1$ convolution layer.

For tactile image inputs, we leverage a frozen V-JEPA 2 ViT-Large~\cite{assran2025v} encoder with 1024-dimensional embeddings. The V-JEPA 2 encoder is pretrained on large-scale video data and produces rich spatiotemporal representations. To adapt these frozen representations to our downstream manipulation tasks while preserving the pretrained knowledge, we introduce learnable residual adapters on top of the ViT encoder.

\paragraph{Residual Adapter Architecture.}
The residual adapter operates on patch-level tokens from the frozen ViT encoder. Given patch tokens $\mathbf{z} \in \mathbb{R}^{N \times D}$ where $N$ is the number of patches and $D = 1024$ is the embedding dimension, the adapter applies:
\begin{equation}
    \mathbf{z}' = \mathbf{z} + \alpha \cdot \text{MLP}(\text{LayerNorm}(\mathbf{z}))
\end{equation}
where $\alpha$ is a learnable scaling parameter initialized to 0.1 for training stability. The MLP consists of 3 layers with hidden dimension 512, GELU activations, and dropout rate 0.1. The adapted patch tokens are then aggregated using attention pooling with a learnable query token:
\begin{equation}
    \mathbf{h}_\tau = \text{AttentionPool}(\mathbf{z}') \in \mathbb{R}^{D}
\end{equation}
This architecture enables task-specific adaptation of the frozen V-JEPA 2 representations with only $\sim$1.5M additional trainable parameters per tactile encoder.

\paragraph{Transformer Architecture.}
The policy transformer follows the ACT architecture with a CVAE-style encoder-decoder structure:
\begin{itemize} [leftmargin=2em, itemsep=0.08em, topsep=0.15em] 
    \item \textbf{CVAE Encoder}: 4-layer transformer encoder with hidden dimension 512, 8 attention heads, feed-forward dimension 3200, and dropout 0.1. The encoder produces a 32-dimensional latent code during training.
    \item \textbf{Action Decoder}: 7-layer transformer decoder with the same hidden dimensions. The decoder takes as input learnable action queries, proprioceptive state embedding, and the latent code.
    \item \textbf{Action Chunk Size}: 45 timesteps, enabling temporal action prediction for smooth execution.
\end{itemize}

\vspace{1mm}
\noindent\textbf{World Model and Tactile Adaptation.}
We employ V-JEPA 2 ViT-L/Vit-G as our tactile world model, pretrained on tactile images from our dataset and frozen during policy training. The pretrained encoder (in the case of ViT-L) produces 1024-dimensional patch embeddings. To enable the policy to refine its draft actions using tactile context while still preserving the pretrained representation, we insert a lightweight residual adapter after the frozen encoder. The adapter processes all patch tokens (not just the CLS token) through a 3-layer bottleneck MLP with GELU activations and dropout ($p{=}0.1$). A learnable residual scale, initialized to 0.1, controls the magnitude of adapter features added to the frozen representations. We aggregate the adapted patches using learned attention pooling, where a single learnable query token attends to all 196 adapted patches via 8-head multi-head attention. This architecture adds only 5.5M trainable parameters (1.8\% overhead) to the 300M frozen ViT-L, enabling efficient task-specific adaptation while retaining the world model's learned dynamics. The adapter and pooling weights are optimized jointly with the policy using AdamW (lr=$1e-5$, weight decay=$1e-4$).

\paragraph{Fusion Strategy.}
Visual features from all cameras (RGB and tactile) are concatenated along the sequence dimension before being processed by the transformer decoder. Specifically, ResNet features are flattened to $H \times W$ tokens per RGB camera, and tactile features contribute 1 pooled token per sensor. Sinusoidal positional embeddings are used for RGB features, and learnable position embeddings are used for tactile and proprioceptive tokens.

\subsection{Training Details}
\label{sec:training}
We use AdamW~\cite{he2016deep} optimizer with the following hyperparameters as shown in Table \ref{tab:hyperparams}.
For RGB images, we apply color jitter augmentation with brightness=0.3, contrast=0.4, saturation=0.5, and hue=0.08 when using diffusion-based policies. Tactile images are only resized to $224 \times 224$ without additional augmentation to preserve contact pattern fidelity.
Actions and proprioceptive states are normalized using per-dimension mean and standard deviation computed from the training set. Images are normalized using ImageNet statistics (mean=[0.485, 0.456, 0.406], std=[0.229, 0.224, 0.225]).
Training is performed on a single NVIDIA GPU. A typical training run of 20,000 steps takes approximately 8--12 hours depending on GPU type and dataset size. We formalize the full DreamTacVLA training sequence in Algorithm \ref{alg:dreamtacvla}.

\begin{table}[h]
\centering
\small
\caption{Complete hyperparameter configuration for DreamTacVLA.}
\begin{tabular}{lc}
\toprule
Hyperparameter & Value \\
\midrule
\multicolumn{2}{c}{\textit{Transformer}} \\
Hidden dimension & 512 \\
Feed-forward dimension & 3200 \\
Encoder layers & 4 \\
Decoder layers & 7 \\
Attention heads & 8 \\
Dropout & 0.1 \\
Latent dimension (CVAE) & 32 \\
Action chunk size & 45 \\
\midrule
\multicolumn{2}{c}{\textit{Visual Encoders}} \\
RGB backbone & ResNet-18 (frozen) \\
RGB input resolution & $640 \times 480$ \\
Tactile encoder & V-JEPA2 ViT-L (frozen) \\
Tactile embedding dim & 1024 \\
Tactile input resolution & $224 \times 224$ \\
\midrule
\multicolumn{2}{c}{\textit{Residual Adapter}} \\
Hidden dimension & 512 \\
Depth (layers) & 3 \\
Dropout & 0.1 \\
Scale init ($\alpha$) & 0.1 \\
Pooling & Attention \\
\midrule
\multicolumn{2}{c}{\textit{HSA Loss}} \\
Temperature ($\kappa$) & 0.07 \\
Feature extractor & ViT-B/16 \\
HSA weight ($\beta_{\mathrm{HSA}}$) & 1.0 \\
\midrule
\multicolumn{2}{c}{\textit{Training}} \\
Optimizer & AdamW \\
Learning rate & $1 \times 10^{-5}$ \\
Backbone learning rate & $1 \times 10^{-5}$ \\
Weight decay & $1 \times 10^{-4}$ \\
Batch size & 16 \\
KL weight ($\beta_{\mathrm{KL}}$) & 10 \\
Training steps & 10,000--20,000 \\
\bottomrule
\end{tabular}

\label{tab:hyperparams}
\end{table}

\subsection{Inference}
\label{sec:inference}

During inference, the CVAE encoder is bypassed and the latent code is set to zero (mean of the prior). Actions are predicted in chunks of 45 timesteps. We use temporal aggregation with exponential weighting to smooth action execution:
\begin{equation}
    a_t = \sum_{k=0}^{K} w_k \cdot a_t^{(k)}
\end{equation}
where $a_t^{(k)}$ is the action at time $t$ predicted $k$ steps ago, and $w_k \propto \exp(-\lambda k)$ are exponentially decaying weights. The control loop runs at 50 Hz to match the data collection frequency.

% \section{Model Architecture and Hyperparameters}
% Multimodal Encoders ($E_{\psi}$)Our system utilizes a CLIP ViT-L/14 backbone for visual and language processing.Language: Task instructions (e.g., "Insert USB") are encoded into language tokens.Vision: Third-person and wrist camera images are patchified into visual tokens.Spatial Alignment: During Stage 1, the Hierarchical Spatial Alignment (HSA) loss pulls corresponding tactile and visual tokens together in a shared latent space using an InfoNCE objective.

% Tactile World Model ($W_{\phi}$) and Think-Dream-ActWorld Model Backbone: We employ V-JEPA2 (ViT-L/G) as the tactile world model. It is pre-trained on large-scale unlabeled tactile sequences and remains frozen during policy training to provide stable physics-based embeddings.Forecasting MLP ($F_{\eta}$): A lightweight forecasting module that predicts the future tactile state $H_{dream}^{(t+N)}$ based on the current state and a draft action.The Refinement Loop: The policy executes in two passes: first generating a draft action $a_{draft}^{(t)}$ with a "null dream" (zero-tensor), and then refining it into $a_{final}^{(t)}$ after integrating the dreamed tactile future.

\section{Hierarchical Spatial Alignment (HSA)}\label{appendix:HSA}

\begin{table}[h]
\centering
\caption{DH parameters for Dobot Nova 2 robot.}
\small
\begin{tabular}{ccccc}
\toprule
Joint & $\theta$ & $d$ (m) & $a$ (m) & $\alpha$ \\
\midrule
1 & $q_0$ & 0.2234 & 0 & $\pi/2$ \\
2 & $q_1 - \pi/2$ & 0 & -0.280 & 0 \\
3 & $q_2$ & 0 & -0.225 & 0 \\
4 & $q_3 - \pi/2$ & 0.1175 & 0 & $\pi/2$ \\
5 & $q_4$ & 0.120 & 0 & $-\pi/2$ \\
6 & $q_5$ & 0.088 & 0 & 0 \\
\bottomrule
\end{tabular}

\label{tab:dh_params}
\end{table}
\subsection{Sensor and Camera Calibration}
\textbf{Extrinsics.} We calibrate the rigid transforms from robot base to each camera frame (third-person and wrist) using a standard hand–eye calibration procedure. Let ${T_c}^b$ denote camera extrinsics in the base frame, and ${T}_{ee}^b$ the end-effector pose from forward kinematics computed using Denavit-Hartenberg (DH) parameters for the Nova 2 robot as shown in Table \ref{tab:dh_params}. The tactile sensor pose $T^{b}_{\mathrm{sensor}}(t)$ is obtained via a fixed transform $T^{ee}_{\mathrm{sensor}}$ mounted on the gripper:
$T^{b}_{\mathrm{sensor}}(t) = {T}_{ee}^b T^{ee}_{\mathrm{sensor}}$, where 
$$ T_{c}^{b}= \begin{bmatrix}
    1.0 & 0.0 & 0.0 & 0.7 \\
    0.0 & -1.0 & 0.0 & -0.49 \\
    0.0 & 0.0 & 1.0 & 1.14 \\
    0.0 & 0.0 & 0.0 & 1.0
    \end{bmatrix}
$$

\textbf{Intrinsics.} For each camera, we use intrinsic matrix K and distortion coefficients from calibration. If images are undistorted online, the projection below assumes the rectified pinhole model. In our case, the value of $K$ is
$$
K = \begin{bmatrix}
    647.0 & 0.0 & 653.0 \\
    0.0 & 644.0 & 364.0 \\
    0.0 & 0.0 & 1.0
\end{bmatrix}
$$

\subsection{Projecting Tactile Sensor to Image Bounding Boxes}
To localize the tactile sensor in the wrist and third-person images for HSA, we project a set of 3D points that approximate the tactile sensor’s visible surface (e.g., 4 corners of a small rectangle / circle sample points) from the sensor frame into each camera:
$\mathbf{u} \sim K\; {}^cT_b\; \mathbf{X}_b,
\quad \mathbf{X}_b = {}^bT_{\mathrm{sensor}}(t)\; \mathbf{X}_{\mathrm{sensor}}.$ We then take the min/max pixel coordinates over the projected points to form bounding boxes $B^{(t)}_w$ and $B^{(t)}_{tp}$. We clip boxes to image bounds and discard frames where (i) the box is fully out of view or (ii) projected depth is negative. We optionally enlarge the boxes by a small margin (e.g., 5–15 pixels) to tolerate calibration noise. If the tactile sensor is occluded in third-person view (common), we keep $B_{tp}$ but down-weight the TPV term in $L_\mathrm{{HSA}}$ or mask it when visibility checks fail.
\subsection{Mapping Bounding Boxes to Patch Tokens}
For CLIP-ViT-large-patch14 inputs of resolution $224\times224$ and patch size 16, the patch grid is $16\times16$. We map bounding box pixels $[x_0,x_1]\times[y_0,y_1]$ to patch indices:
$$
i \in \left[\left\lfloor \frac{x_0}{14}\right\rfloor,\left\lfloor \frac{x_1}{14}\right\rfloor\right],\quad
j \in \left[\left\lfloor \frac{y_0}{14}\right\rfloor,\left\lfloor \frac{y_1}{14}\right\rfloor\right].
$$
Tokens whose patch centers fall inside the box are selected and mean-pooled to form $h_w$ and $h_{tp}$. The tactile embedding $h_\tau$ is mean-pooled from tactile tokens (or from the pooled tactile representation if a pooling head is used). 
\subsection{HSA Loss: Negatives, Temperature, and Weighting}
We compute InfoNCE losses for tactile–wrist and tactile–TPV alignment:
$$
\mathcal{L}_{\mathrm{HSA}}=\mathcal{L}_{\mathrm{HSA\mathrm{-}W}} + \mathcal{L}_{\mathrm{HSA\mathrm{-}TP}}.
$$
\textbf{Negatives.} We use a mix of: in-image negatives: patches outside the bounding box (hard negatives), in-batch negatives: patches from other samples in the batch (diverse negatives).

\textbf{Temperature.} We set $\kappa$ to a fixed value (0.07) and keep it constant across training.

\textbf{Visibility-aware weighting.} If TPV visibility is unreliable, we apply
$\mathcal{L}_{\mathrm{HSA}} = \mathcal{L}_{\mathrm{HSA\text{-}W}} + \alpha(t)\, \mathcal{L}_{\mathrm{HSA\text{-}TP}}$, where $\alpha(t)\in[0,1]$ is set based on projection validity.

\section{Tactile World Model Pretraining and Forecasting Loss}\label{appendix:world_model}

To train our tactile world model, we adopt a self-supervised latent-prediction approach inspired by the \textbf{V-JEPA} (Video Joint-Embedding Predictive Architecture) framework. This approach shifts the learning objective from raw sensor reconstruction to the prediction of abstract latent representations.

\subsection{Latent Masked Forecasting via Teacher Forcing}

The pretraining phase utilizes a \textit{teacher-student} architecture to facilitate stable feature extraction from tactile sequences. This process is defined by two primary components:

\begin{itemize} [leftmargin=2em, itemsep=0.08em, topsep=0.15em]
    \item \textbf{Teacher Encoder:} The teacher receives the complete, unmasked tactile sequence and generates target embeddings. To maintain a stable training target and prevent representation collapse, the teacher's weights are updated using an Exponential Moving Average (EMA) of the student's parameters rather than through direct backpropagation.
    \item \textbf{Student Predictor:} The student is provided with a temporally masked spatial version of the tactile sequence. Its objective is to predict the latent representations of the missing patches, conditioned on the available context and the specific positional encodings of the masked regions.
\end{itemize}

\subsection{Forecasting Loss Function}

The world model is optimized by minimizing the $L_2$ distance between the student's predicted embeddings and the teacher's ground-truth embeddings in the latent space. Critically, the loss is calculated only over the masked indices $M$. 

The forecasting loss $\mathcal{L}$ is formulated as:
\begin{equation}
\mathcal{L} = \sum_{t \in M} \| z_{\mathrm{teacher}}(t) - \hat{z}_{\mathrm{student}}(t) \|^2_2,
\end{equation}
where $z_{\mathrm{teacher}}(t)$ represents the latent target produced by the EMA-updated teacher, and $\hat{z}_{\mathrm{student}}(t)$ represents the student's prediction for the corresponding masked segment. By training in this latent space, the model learns to capture the underlying physics of tactile interaction—such as shear forces and surface geometry—while remaining robust to the inherent sensor noise found in raw tactile data.

\subsection{Pretraining Data and Temporal Sampling}
We pretrain the tactile world model on unlabeled tactile sequences extracted from both simulation and real demonstrations. We sample short clips of length T with stride s, and randomly choose a prediction horizon $N\in \{1,\dots,N_{\max}\}$. This yields pairs $(I_\tau^{(t)}, I_\tau^{(t+N)})$ to encourage multi-step predictive structure.

\section{Evaluation Protocol}\label{appendix:evaluation}
To evaluate the efficacy of our framework, we conducted extensive experiments on a manipulation platform. Our evaluation focuses on the system's ability to perform high-precision tasks where visual and tactile integration is critical.

The hardware setup consists of a Dobot XTrainer platform. The sensing suite are two wrist-mounted cameras provide localized views of the grippers, while a single overhead camera captures the global workspace and a Digit sensor is integrated into the end-effector, providing high-resolution tactile feedback of the contact geometry.

\begin{figure}
    \centering
    \includegraphics[width=0.98\linewidth]{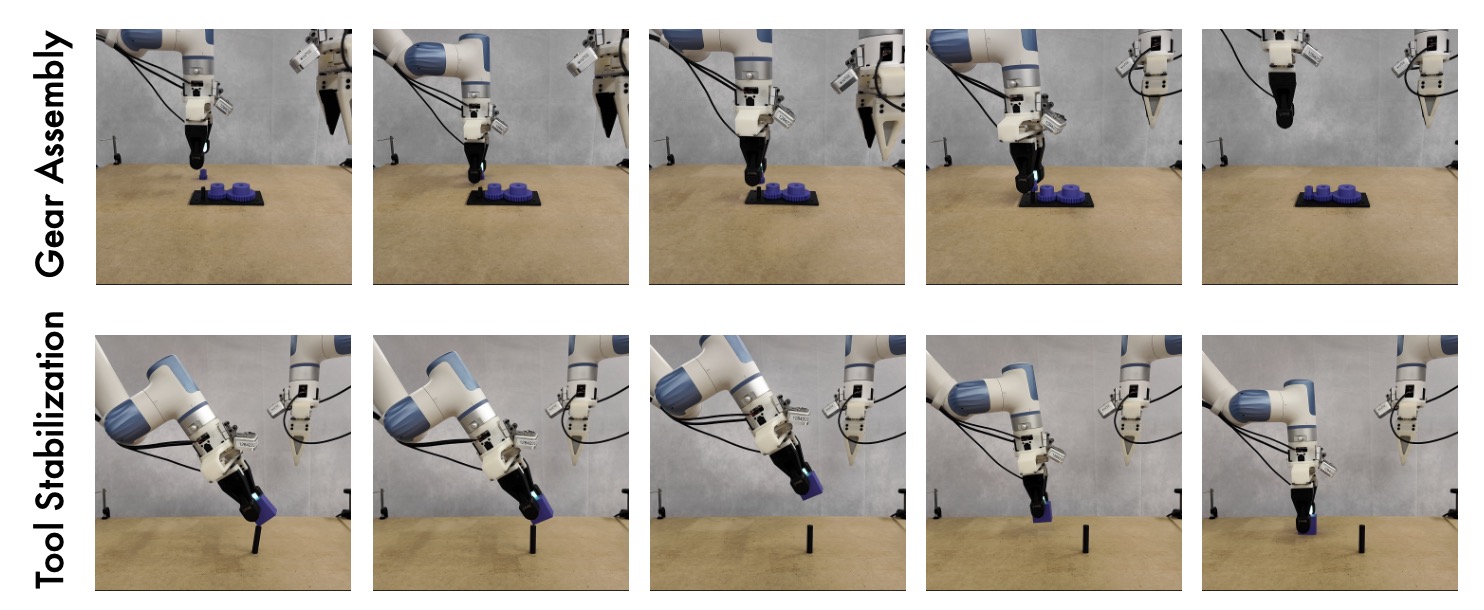}
    \caption{Keyframes of the gear assembly and tool stabilization task workflow.}
    \label{fig:keyframes_two_tasks}
\end{figure}

During evaluation, objects are placed at fixed initial positions. This protocol allows us to isolate and measure the model's performance on \textit{precision-based} execution and fine-grained adjustments, effectively decoupling the control performance from variables associated with open-world localization. To ensure the statistical significance of our results and assess the reliability of the framework, we conducted over 100 trials for each task. This extensive testing allows us to accurately characterize the success rate and error distributions of the system. We show the rollout sequences of gear assembly and tool stabilization task in Figure~\ref{fig:keyframes_two_tasks}. A failure case for Peg In Hole is also shown in Figure~\ref{fig:keyframes_failure_case}. All the videos and source code can be found in supplementary materials.

\begin{figure}
    \centering
    \includegraphics[width=0.9\linewidth]{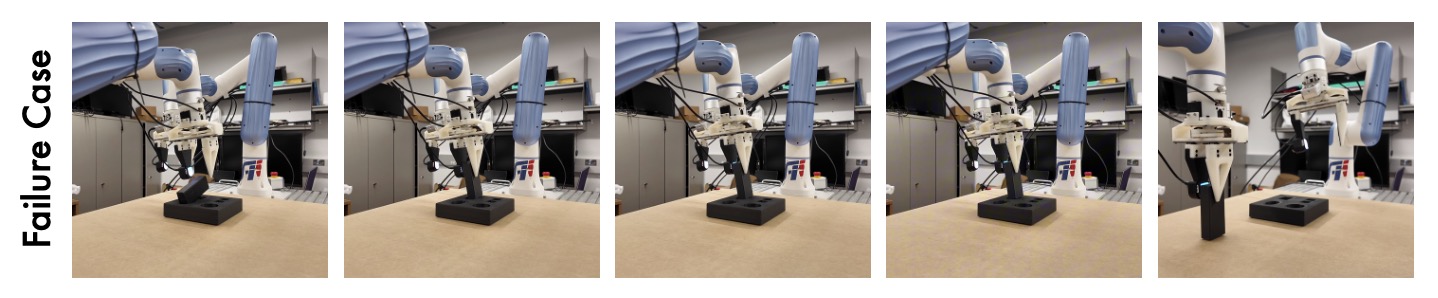}
    \caption{Keyframes of one example failure case. }
    \label{fig:keyframes_failure_case}
\end{figure}
%evaluation rollout figures

\section{Extended Related Work}
\label{appendix:extended_related_work}
\subsection{Vision-Language-Action (VLA) Models}
\label{appRel1}
Vision-Language-Action (VLA) models unify perception, language understanding, and action generation into a single policy for general-purpose robot control. Large-scale systems such as RT-1~\cite{Brohan-RSS-23} and RT-2~\cite{pmlr-v229-zitkovich23a} demonstrate that scaling data and model capacity enables strong cross-task and cross-embodiment generalization. Subsequent work extends this paradigm across embodiments, datasets, and action spaces, including Open X-Embodiment~\cite{o2024open}, OpenVLA~\cite{pmlr-v270-kim25c}, Octo~\cite{Ghosh-RSS-24}, RDT-1B~\cite{ICLR2025_49f80e4d}, $\pi_0$~\cite{black2024pi_0}, $\pi_{0.5}$~\cite{pmlr-v305-black25a}, and GR00T N1~\cite{bjorck2025gr00t}.

Beyond scale, architectural and training refinements further improve VLA effectiveness, including diffusion-based action generation~\cite{chi2025diffusion} and language-conditioned visuomotor learning for compositional generalization~\cite{Zhao-RSS-23}. Modular designs such as CogACT~\cite{li2024cogact} decouple high-level reasoning from low-level control, and optimized adaptation strategies like OpenVLA-OFT~\cite{kim2025fine} show that carefully designed fine-tuning can significantly boost real-robot performance and efficiency.

Despite these advances, most VLA models remain vision-centric and struggle in contact-rich manipulation, where RGB observations alone cannot resolve contact state, friction, or incipient slip. Recent work such as FuSe~\cite{11127987} explores post-hoc adaptation with tactile and audio signals, but tactile is often treated as auxiliary rather than being integrated into perception, prediction, and action generation.

\subsection{Spatial Grounding for Robotics}
\label{appRel2}
Spatial grounding improves visuomotor reasoning by incorporating geometric structure beyond RGB observations. Recent VLA variants introduce explicit 3D priors, including egocentric position encodings in SpatialVLA~\cite{QuD-RSS-25} and point cloud grounding in PointVLA~\cite{11346992}, which improve generalization in geometry-sensitive manipulation. Complementary approaches model spatial structure implicitly: TraceVLA~\cite{ICLR2025_8667f264} encodes spatiotemporal interaction traces, while Evo-0~\cite{lin2025evo} injects implicit 3D structure into the visual backbone via architectural and training biases. These methods primarily reason at the level of object pose and scene layout, leaving contact-level geometry unmodeled.

\subsection{Tactile Grounding for Manipulation}
\label{appRel3}
Tactile grounding provides contact information unavailable to vision, including contact geometry, friction, and slip, which are critical for contact-rich manipulation. Early approaches rely on low-dimensional force/torque sensing, which supports contact detection and impedance control but discards spatial structure, leading to ambiguous contact states.

Beyond force-based sensing, several works integrate tactile signals into learned policies. See, Hear, and Feel~\cite{pmlr-v205-li23c} studies structured multimodal fusion, while more recent VLA-style approaches incorporate tactile inputs directly. MLA~\cite{liu2025mla}, Tactile-VLA~\cite{huang2025tactilevla}, OmniVTLA~\cite{cheng2025omnivtla}, and RDP~\cite{XueH-RSS-25} demonstrate improved robustness via vision–tactile fusion and reactive tactile feedback, but typically operate on temporally sparse or spatially compressed tactile representations.

In parallel, tactile representation learning focuses on transferable visuotactile embeddings decoupled from control, including Binding Touch~\cite{yang2024binding}, TVL~\cite{pmlr-v235-fu24b}, Sparsh~\cite{pmlr-v270-higuera25a}, AnySkin~\cite{bhirangi2025anyskin}, and T3~\cite{pmlr-v270-zhao25c}. To enrich contact signals, ViTacGen~\cite{11204497} synthesizes visuotactile images from RGB inputs, but remains constrained by visual observability.

In contrast, vision-based tactile sensors such as GelSight~\cite{s17122762} and DIGIT~\cite{lambeta2020digit} provide dense micro-vision measurements of surface deformation, encoding texture, geometry, and shear-induced slip, which are essential for modeling fine-grained contact interactions~\cite{}. However, existing approaches largely treat tactile sensing as an auxiliary input or a standalone representation, without tightly integrating tactile prediction into sequential decision making.

\subsection{Predictive World Models in Robotics}
\label{appRel4}
Predictive world models learn latent dynamics that enable anticipation of future observations for planning and control. Early work such as World Models~\cite{ha2018world} and Dreamer~\cite{hafner2023mastering} demonstrate imagination-based control via recurrent latent dynamics. Recent large-scale pretraining methods, including R3M~\cite{pmlr-v205-nair23a} and V-JEPA-style approaches~\cite{assran2025v}, learn structured predictive representations that transfer effectively to downstream robotic tasks.

Several VLA architectures integrate predictive modeling directly into policy learning by conditioning actions on latent rollouts, including DreamVLA~\cite{NEURIPS2025_22d4f952} and WorldVLA~\cite{cen2025worldvla}. In the tactile domain, ViTacFormer~\cite{heng2025vitacformer} explores autoregressive visuotactile prediction, but remains focused on representation learning rather than action-conditioned rollout.

Overall, existing world-model approaches predominantly operate on visual observations or abstract latents, and lack explicit mechanisms for predicting fine-grained contact evolution, motivating extensions toward tactile-aware predictive modeling.

% \subsection{Unified Positioning and Cross-Cutting Comparisons}